\documentclass[]{uiuc_paper}
\pdfoutput=1
\usepackage[toc,page,header]{appendix}
\usepackage{cleveref}
\usepackage{subcaption}
\usepackage{booktabs}
\usepackage{tabularx}

\usepackage{latexsym}
\usepackage{url}
\usepackage{amssymb}
\usepackage{amsmath}
\usepackage{amsfonts}
\usepackage[utf8]{inputenc}
\usepackage{booktabs}
\usepackage{pifont}
\usepackage{multirow}
\usepackage{makecell}
\usepackage{paralist}
\usepackage{xspace}
\usepackage{adjustbox}

\usepackage{colortbl}
\usepackage{algorithm}
\usepackage{algpseudocode}
\definecolor{oursbg}{HTML}{E3EFF9}
\definecolor{gain}{HTML}{2D6A99}
\definecolor{cited}{HTML}{677185}
\newcommand{\ours}{\rowcolor{oursbg}}
\newcommand{\gainrow}[1]{\quad{\footnotesize\emph{#1}}}
\crefname{section}{Sec.}{Secs.}\Crefname{section}{Sec.}{Secs.}
\crefname{subsection}{Sec.}{Secs.}\Crefname{subsection}{Sec.}{Secs.}
\crefname{subsubsection}{Sec.}{Secs.}\Crefname{subsubsection}{Sec.}{Secs.}
\crefname{appendix}{App.}{Apps.}\Crefname{appendix}{App.}{Apps.}
\crefname{subappendix}{App.}{Apps.}\Crefname{subappendix}{App.}{Apps.}
\crefname{subsubappendix}{App.}{Apps.}\Crefname{subsubappendix}{App.}{Apps.}
\crefname{figure}{Fig.}{Figs.}\Crefname{figure}{Fig.}{Figs.}
\crefname{table}{Tab.}{Tabs.}\Crefname{table}{Tab.}{Tabs.}
\crefname{equation}{Eq.}{Eqs.}\Crefname{equation}{Eq.}{Eqs.}
\crefname{algorithm}{Alg.}{Algs.}\Crefname{algorithm}{Alg.}{Algs.}
\creflabelformat{equation}{#2#1#3}
\newcommand{\method}{Bel\xspace}

\hypersetup{
  bookmarksnumbered=true,
  pdftitle={Bel: Step-Level Policy Improvement of Visual Agents Without Training},
  pdfauthor={Jiajun Fan, Ziwen Wang, Zhaoyu Zhu, Changyi Yang, Neil He, Chaoran Cheng, Ge Liu}}

\title{Bel: Step-Level Policy Improvement\\of Visual Agents Without Training}

\author[1]{Jiajun Fan}
\author[1,2]{Ziwen Wang}
\author[1]{Zhaoyu Zhu}
\author[3]{Changyi Yang}
\author[1,4]{Neil He}
\author[1]{Chaoran Cheng}
\author[1]{Ge Liu}

\affiliation[1]{\affillogo{\includegraphics{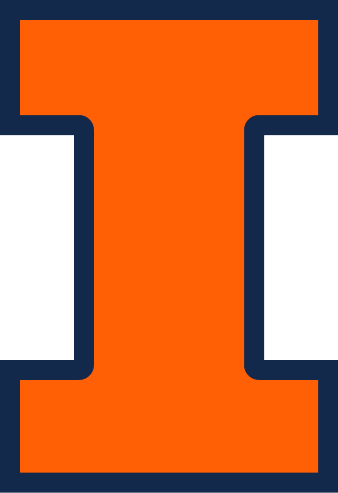}}{UIUC}}
\affiliation[2]{\affillogo{\includegraphics{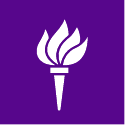}}{NYU}}
\affiliation[3]{\affillogo{\includegraphics{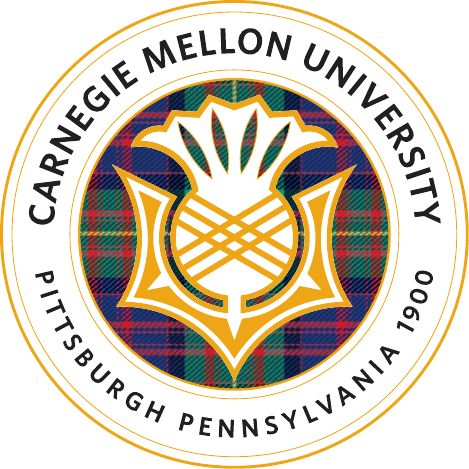}}{CMU}}
\affiliation[4]{\affillogo{\includegraphics{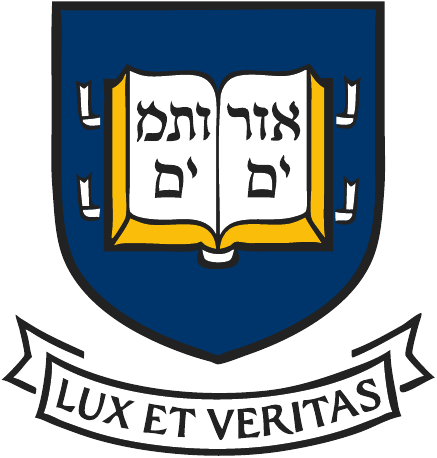}}{Yale}}

\abstract{A visual agent acts on what it sees: it taps and types on a screen, or points on a camera image to show a robot where to act.
One wrong step can fail its task, yet the right action is often among its own samples, passed over by its greedy decode.
Choosing it is a visual judgment, since an action is best judged where it acts, not by its text or coordinates alone: as text, a tap is two numbers; drawn on the screen, it is the \emph{Post} button of an empty draft.
We find that this is decisive: given the screenshot and a tap's coordinates, a decision model chooses among GUI-R1-3B's samples significantly worse on average than when the tap is drawn where it lands.
\method turns this finding into a method that trains nothing and needs only a frozen policy's samples and an image on which each action can be drawn.
It \emph{marks} each candidate on its own copy of the image it acts on, \emph{asks} a decision model for the probability that executing it is the right decision, so that every kind of action shares one scale, and \emph{acts} on another candidate only when greedy scores below a threshold and trails it by a margin.
On recorded GUI steps, for each of three policies, \method significantly beats greedy decoding on mobile, web and desktop screens alike and, on the same candidates, gives each a significantly higher mean step success than voting, CATTS, the trained GUI-Critic-R1 and two prompted Qwen3-VL judges shown the same marks; it also significantly improves the much larger Qwen3-VL-235B-A22B on every step-level test set.
With GUI-R1-3B's thresholds unchanged, \method significantly raises live AndroidWorld and MiniWoB++ task success for two other policies and robot-pointing success for three policies on each of two groups of benchmarks.
Judging each action where it acts thus gives step-level policy improvement of visual agents without training, with significant gains for GUI agents and robot pointing.
}

\date{\today}

\begin{document}

\maketitle

\begin{figure}[t]
\centering
\includegraphics[width=\linewidth]{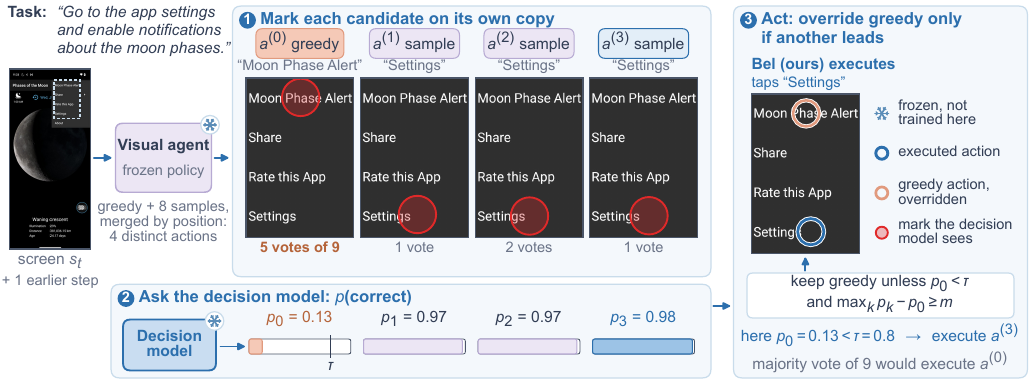}
\caption{\textbf{\method marks, asks, and acts.} The frozen policy proposes its greedy action and eight samples; each distinct candidate is drawn on its own copy of the screenshot, the decision model returns $p(\text{correct})$ for each, and greedy is replaced only when it scores below $\tau$ and another candidate leads it by $m$. A real AndroidControl-High step of GUI-R1-3B (rule: \cref{sec:app_cases}): voting would execute ``Moon Phase Alert'', \method executes ``Settings''.}
\label{fig:method}
\end{figure}

\begin{figure}[t]
\centering
\includegraphics[width=\linewidth]{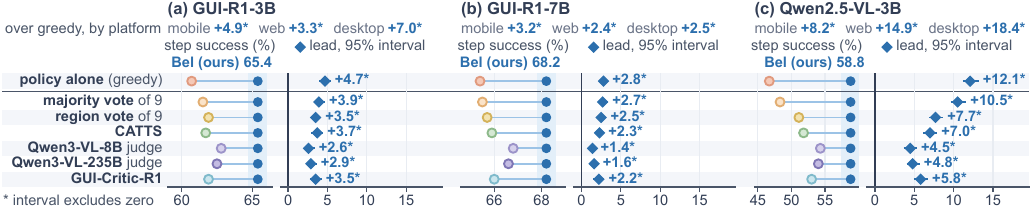}
\caption{\textbf{For every policy, \method's six-set mean step success is significantly above every selector of \cref{tab:main}, and its gain over greedy decoding is significant on mobile, web and desktop screens alike.} Mean step success (higher is better) and \method's lead over each selector with its paired 95\% interval (star: excludes zero); above each panel: gain over greedy per platform; leads and gains: of rounded values.}
\label{fig:hero}
\end{figure}

\section{Introduction}
\label{sec:intro}

Visual agents act on what they see, one step at a time: GUI agents on phones, in browsers and on desktops \cite{rawles2023aitw,li2024androidcontrol,cheng2024seeclick,qin2025uitars}, and models that point on a robot's camera image to show it where to act \cite{yuan2024robopoint,zhou2025roborefer}; most are vision--language models, fine-tuned \cite{cheng2024seeclick,qin2025uitars} or trained with reinforcement learning \cite{luo2025guir1,lu2025uir1}.
One wrong step can fail the whole task, yet the right action is often not missing but unchosen: with eight samples, GUI-R1-3B has a right candidate on far more steps than its greedy action gets right (\cref{fig:analysis}a).
Picking it would improve any frozen agent without touching its weights, but the choice is a visual judgment, and it poses two problems.
First, a point's coordinates do not show what it would hit, even next to the image: as text, a tap is two numbers; drawn on the screen, it is the \emph{Post} button of an empty draft.
Second, candidates differ in kind (a tap, a typed string, a scroll, a decision to stop) and must be compared on one scale.

Existing selectors solve them only in part.
Majority voting \cite{wang2023selfconsistency} and its region variant for clicks \cite{du2026guirc} never look at the screen, so a right action few samples propose loses to a wrong one many propose; CATTS \cite{lee2026catts} lets the policy choose only when the vote is contested.
Trained GUI critics \cite{wanyan2025guicritic,wang2026gaia} and GUI-Actor's click verifier \cite{wu2025guiactor} look at the screen but are built for the platforms and action spaces of their labels, and the verifier judges only where a click lands.
Prompted judges take a verdict \cite{zheng2023judging,pan2024autonomous}, or the probability of \texttt{Yes}, from a general model; given the same marks, open ones select worse than \method (\cref{tab:main}). Process reward agents \cite{xiong2025guipra} run an agent per candidate.

We solve both problems with \method (for \emph{belief}), which judges each candidate where it acts (\cref{fig:method}), using Jev \cite{typesafe2026jev}, a decision model that reads images.
\emph{Mark} solves the first problem: every distinct candidate gets its own copy of the image it acts on (a screenshot or a robot's camera image), with any point drawn where it lands, so each is judged alone by what it would hit.
\emph{Ask} solves the second: the decision model returns the probability that executing the candidate is the right decision, so every kind of action lands on one scale.
\emph{Act} protects the agent's own choice: greedy is kept unless it scores below a threshold and another candidate leads it by a margin.
\method trains nothing; its two thresholds are chosen per policy on a development set, and GUI-R1-3B's are reused unchanged in the live and robot runs.
In summary, our contributions are:
\begin{itemize}
\item \textbf{A finding: an action must be judged where it acts, not by its coordinates.} Given the screenshot and the tap's coordinates, the decision model recovers significantly less of the gain on average; drawing each tap where it lands gives the full gain (GUI-R1-3B, \cref{sec:ablation}). This is the principle behind \method.
\item \textbf{Mark, ask, act: a method for any frozen visual agent.} It needs only the policy's samples and an image on which each action can be drawn, and keeps the gain of always taking the highest score while changing far fewer steps (GUI-R1-3B, \cref{sec:ablation}). Two thresholds are its only per-policy setting, and every pair of the grid beats greedy decoding for each policy of \cref{tab:main} (\cref{sec:app_robust}).
\item \textbf{It repairs wrong decisions, not only off-target taps.} Nearly all repaired steps had the wrong action type---on GUI-Odyssey often a premature stop---or the wrong element, including steps on which every other selector errs (GUI-R1-3B, \cref{sec:qual}).
\item \textbf{Step-level policy improvement of visual agents without training, with significant gains for GUI agents and robot pointing.} For every policy of \cref{tab:main}, \method significantly raises step success (on mobile, web and desktop screens alike) and grounding accuracy, and is significantly ahead of voting, CATTS, GUI-Critic-R1 and prompted Qwen3-VL judges shown the same marks (\cref{tab:main}, \cref{fig:hero}); on Multimodal-Mind2Web it is significantly ahead of every other selector of \cref{tab:m2w}, including ICM-r2, trained on the benchmarks of three of our test sets. It also significantly improves the larger Qwen3-VL-235B-A22B on every step-level test set. With GUI-R1-3B's thresholds unchanged, it significantly raises live AndroidWorld and MiniWoB++ success for two other policies and robot-pointing success for three policies on each of two benchmark groups (\cref{sec:exp}).
\end{itemize}

\section{Method}
\label{sec:method}

\method improves a frozen visual agent step by step in three moves (\cref{fig:method}): it marks each candidate where it acts, asks a decision model how likely each is to be right, and acts on another only when it leads greedy.

\subsection{Problem Formulation}
\label{sec:formulation}
A visual agent receives an instruction $x$ and, at step $t$, an image $s_t$ (a screenshot or a robot's camera image) and the history $h_t$ of earlier steps.
Its policy $\pi$ emits an action $a_t$: an action type (click, long press, type, scroll, open an app, press a key, \dots) with its arguments (a point, a string, a direction), or, for robot pointing, a point.
We assume only sampling access to a frozen policy.
At each step we take the greedy action $a^{(0)}$, the one the agent would execute anyway, and $N$ sampled actions $a^{(1)},\dots,a^{(N)}$ from the same prompt.
To score each distinct action once, candidates of the same type are merged when their points fall in the same cell of a $100{\times}100$ grid over the image and their strings or directions match.
This leaves $K \le N{+}1$ distinct candidates $a^{(0)},\dots,a^{(K-1)}$, greedy first, then in order of first appearance; \method's new policy $\pi'$ executes one of them.

\subsection{Mark: Drawing Each Candidate Where It Acts}
\label{sec:mark}
Two coordinates do not show what a point would hit, so an action with a point is drawn on its own copy $\tilde{s}_k$ of $s_t$ as a translucent red circle with a radius of about 2.5\% of the longer side; it is also written as text, with its coordinates in the shown image (\cref{sec:app_prompts}).
Actions without a point (type, scroll, open app, key presses) are given as text on the unmarked screenshot.
The $K$ queries (instruction, history, marked image, action) run in parallel.

\subsection{Ask: One Probability Scale for Every Action}
\label{sec:ask}
Each query poses six questions of two answers each to Jev \cite{typesafe2026jev}, a decision model that accepts images and returns a probability for every answer.
We call Jev through its public API, train nothing and give it no labels (release, dependence on the screen and repeatability: \cref{sec:app_prompts}).
The score of candidate $k$, $p_k = p(\text{correct} \mid x, h_t, \tilde{s}_k, a^{(k)})$, is the probability of the verdict that the decision is correct and should be executed.
Judged alone, each candidate gets an absolute score, not a rank: a tap, a typed string and a stop are scored on the same footing, by how likely each is to be right in this image, so a single pair of thresholds is used for every kind of action and every GUI test set (\cref{sec:app_thresholds}).
The same call answers five further questions (wording: \cref{sec:app_prompts}), such as whether the mark lands on the intended element; none enters the selection, and three are analysed in \cref{sec:qual}.

\subsection{Act: When to Override Greedy}
\label{sec:act}
Let $k^\star = \arg\max_k p_k$.
The new policy executes
\begin{equation}
\pi'(x, h_t, s_t) =
\begin{cases}
a^{(k^\star)} & \text{if } p_0 < \tau \text{ and } p_{k^\star} - p_0 \ge m,\\
a^{(0)} & \text{otherwise,}
\end{cases}
\label{eq:gate}
\end{equation}
where $\tau$ is the score below which the greedy action may be replaced, and $m$ is the margin another candidate needs; $\tau{=}1.01$ removes the bound, and ties for $k^\star$ go to the earlier candidate (pseudocode: \cref{sec:app_algorithm}).
Overriding a right greedy action loses the step, so $\pi'$ departs from greedy only where another candidate leads it by $m$ and, for each policy of \cref{tab:main} (development-set $\tau \le 0.8$), on every held-out test set only where greedy itself scores below $\tau$; this keeps the gain of always taking the highest score while changing far fewer steps (GUI-R1-3B, \cref{sec:ablation}).

\subsection{Choosing the Two Thresholds}
\label{sec:thresholds}
The two thresholds are \method's only per-policy setting. They are chosen to maximise step success over a small grid (\cref{sec:app_algorithm}) on AndroidControl-High, our development set, and applied unchanged to every other GUI test set; each offline policy gets its own setting, and the live and robot runs keep GUI-R1-3B's (\cref{sec:app_thresholds}; robustness: \cref{sec:app_robust}).
On the development set itself we report two-fold cross-fitting: its tasks (all steps that share an instruction) are split into two folds, the setting that maximises step success on one fold is applied to the other, and vice versa, so no step chooses its own setting.
Nothing else is set per policy: the score, the questions, the mark and the merging grid are the same everywhere (type, select and swipe: \cref{sec:app_m2w,sec:app_online}). The questions were written for the step-judging study of \cref{sec:app_scorers}, independently of step selection, and which answer is the score was fixed on the development set.

\section{Experiments}
\label{sec:exp}

\subsection{Experimental Setup}
\label{sec:setup}

\paragraph{Eight test sets cover phone, tablet, web and desktop screens.}
We test GUI agents first, offline and live, then robot pointing (\cref{sec:embodied}).
We follow the offline step-level protocol of GUI-R1 \cite{luo2025guir1} on six test sets: AndroidControl \cite{li2024androidcontrol} with high-level (AC-High) and step-level instructions (AC-Low), GUI-Odyssey \cite{lu2024guiodyssey}, GUI-Act-Web \cite{chen2024guicourse}, and OmniAct-Web and OmniAct-Desktop \cite{kapoor2024omniact}.
At every step the agent receives the instruction, any history the benchmark gives, and the screenshot, and predicts one action.
We report step success with GUI-R1's criteria: a point is right within 14\% of the screen of the ground-truth point, and a typed, scrolled or opened argument when its token F1 with the ground truth is at least 0.5, whatever the action type (requiring it too enlarges each policy's mean gain, \cref{sec:app_perset}).
It is computed over the steps whose ground truth has a click target or an argument; unlike GUI-R1's script, we keep in the denominator an action category with no right step (\cref{sec:app_full}).
We also measure grounding on ScreenSpot \cite{cheng2024seeclick} and ScreenSpot-Pro \cite{li2025screenspotpro} with GUI-R1's grounding prompt; a click must lie inside the target element.
AC-High is our development set; all other test sets are held out, though AC-Low shares its screens.
All selectors are compared on fixed evaluation subsets: 2,000 random steps of each of AC-High, AC-Low and GUI-Odyssey, and all five other sets; repaired steps and whole AC-High tasks use the complete test sets.
A difference is significant if its paired 95\% bootstrap interval over tasks (an AC-Low step counts in its AC-High task; a grounding or pointing item is its own task) excludes zero.

\paragraph{Frozen policies give a greedy action and eight samples per step.}
The main policy is the released GUI-R1-3B \cite{luo2025guir1}, a vision--language model trained with reinforcement learning for GUI actions, run with GUI-R1's own prompts and scripts.
To test whether the gain depends on the policy, we repeat the comparison with the larger GUI-R1-7B and with Qwen2.5-VL-3B-Instruct \cite{bai2025qwen25vl}, the general-purpose model from which GUI-R1-3B is trained, run with the same prompts; each policy's thresholds are chosen on its own development-set candidates.
The much larger Qwen3-VL-235B-A22B \cite{qwen3vl}, served through an API, is run with the same prompts (\cref{sec:app_q235}).
At each step we take the greedy action (temperature 0, the official protocol) and $N{=}8$ samples at temperature 1.0.

\begin{figure}[t]
\centering
\includegraphics[width=0.6\linewidth]{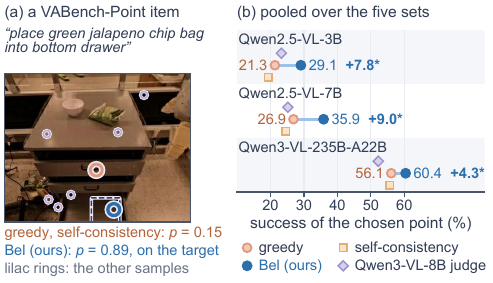}
\caption{\textbf{Without retuning, \method picks significantly better robot points than greedy decoding, self-consistency and the judge for three policies.} (a) A VABench-Point item (Qwen2.5-VL-7B). (b) Success over the five pointing sets; over Part-Afford and Point-It-Out the gain is significant for Qwen2.5-VL-3B, -7B and Qwen3-VL-8B (\cref{sec:app_embodied}); label: gain over greedy (star: 95\% interval excludes zero). Higher is better.}
\label{fig:embodied}
\end{figure}

\paragraph{Every selector chooses among the same candidates.}
\emph{Majority vote} picks the most frequent candidate, with ties broken towards greedy; \emph{region vote} adapts the rule of GUI-RC \cite{du2026guirc}, executing the click nearest the centre of the area most clicks cover (other actions: exact match; \cref{sec:app_prompts}).
\emph{GUI-Critic-R1} \cite{wanyan2025guicritic}, a 7B critic trained, partly on GUI-Odyssey, to diagnose a GUI operation before it is executed, receives its official prompt with the instruction, history, marked screenshot and action; its score is the probability of \texttt{Correct}.
The \emph{Qwen3-VL-8B judge} is Qwen3-VL-8B-Instruct \cite{qwen3vl} asked whether the marked action is the right next step, scored by the probability of \texttt{Yes}; the \emph{Qwen3-VL-235B judge} asks Qwen3-VL-235B-A22B the same question, but for a probability from 0 to 100.
These three use the override rule of \cref{eq:gate} with their own thresholds chosen on the development set.
\emph{CATTS} \cite{lee2026catts} executes the majority vote unless the vote is contested, and otherwise lets the policy itself pick one of the distinct candidates, shown with their vote counts; its threshold is chosen on the development set (\cref{sec:app_thresholds}).
\Cref{tab:m2w} adds ICM-r2 \cite{wang2026gaia}, a Qwen2.5-VL-7B critic fine-tuned on AndroidControl and GUI-Odyssey actions, and GUI-Actor's click verifier \cite{wu2025guiactor}; \cref{sec:app_selectors} adds a prompted Claude Sonnet 4.6 judge (\cref{sec:app_prompts}).

\subsection{Main Results}
\label{sec:main}

\begin{table}[t]
\centering\footnotesize
\caption{\textbf{For every policy, \method is ahead of every selector in this table on all three measures, significantly in every column.} Step success (\%, mean over the six GUI test sets) and grounding accuracy (\%); same candidates for every selector, thresholds chosen on AC-High. Bold: best; underline: second; gain: of the printed values; star: its paired 95\% interval excludes zero; intervals of every lead: \cref{sec:app_screenspot}.}
\label{tab:main}
\setlength{\tabcolsep}{3.5pt}
\begin{tabular}{lccccccccc}
\toprule
\emph{higher is better} & \multicolumn{3}{c}{Policy: GUI-R1-3B} & \multicolumn{3}{c}{Policy: GUI-R1-7B} & \multicolumn{3}{c}{Policy: Qwen2.5-VL-3B} \\
\cmidrule(lr){2-4}\cmidrule(lr){5-7}\cmidrule(lr){8-10}
Selector & \shortstack{\vphantom{dy}Step\\\vphantom{dy}succ.} & ScreenSpot & SS-Pro & \shortstack{\vphantom{dy}Step\\\vphantom{dy}succ.} & ScreenSpot & SS-Pro & \shortstack{\vphantom{dy}Step\\\vphantom{dy}succ.} & ScreenSpot & SS-Pro \\
\midrule
Policy alone (greedy) & 60.7 & 81.7 & 28.5 & 65.4 & 85.8 & 31.1 & 46.7 & 64.3 & 13.8 \\
\midrule
Majority vote of 9 & 61.5 & 84.3 & \underline{30.5} & 65.5 & 87.1 & \underline{31.7} & 48.3 & 66.6 & 14.1 \\
Region vote of 9 \cite{du2026guirc} & 61.9 & 84.0 & \underline{30.5} & 65.7 & 86.9 & 31.6 & 51.1 & \underline{70.4} & 13.4 \\
CATTS \cite{lee2026catts} & 61.7 & 83.5 & 28.8 & 65.9 & 86.6 & 31.6 & 51.8 & 62.2 & 11.8 \\
Qwen3-VL-8B judge & \underline{62.8} & \underline{84.5} & 29.7 & \underline{66.8} & 87.1 & 31.3 & \underline{54.3} & 67.4 & 13.1 \\
Qwen3-VL-235B judge & 62.5 & 84.1 & 29.5 & 66.6 & \underline{87.3} & 31.6 & 54.0 & 69.5 & \underline{15.0} \\
GUI-Critic-R1 \cite{wanyan2025guicritic} & 61.9 & 82.1 & 27.9 & 66.0 & 85.9 & 30.4 & 53.0 & 66.9 & 14.0 \\
\ours \method (ours) & \textbf{65.4} & \textbf{88.1} & \textbf{33.7} & \textbf{68.2} & \textbf{89.2} & \textbf{33.8} & \textbf{58.8} & \textbf{79.4} & \textbf{17.7} \\
\ours \gainrow{gain over the policy alone} & {\scriptsize\textcolor{gain}{+4.7\rlap{*}}} & {\scriptsize\textcolor{gain}{+6.4\rlap{*}}} & {\scriptsize\textcolor{gain}{+5.2\rlap{*}}} & {\scriptsize\textcolor{gain}{+2.8\rlap{*}}} & {\scriptsize\textcolor{gain}{+3.4\rlap{*}}} & {\scriptsize\textcolor{gain}{+2.7\rlap{*}}} & {\scriptsize\textcolor{gain}{+12.1\rlap{*}}} & {\scriptsize\textcolor{gain}{+15.1\rlap{*}}} & {\scriptsize\textcolor{gain}{+3.9\rlap{*}}} \\
\bottomrule
\end{tabular}
\end{table}

\begin{figure}[t]
\begin{minipage}[c]{0.485\linewidth}
\centering\footnotesize
\captionof{table}{\textbf{On Multimodal-Mind2Web, \method is the best selector for every policy.} Step success (\%), mean over the three test splits; ICM-r2 under the better of our rule and GAIA's \cite{wang2026gaia}. Bold: best; underline: second; star: \method is significantly above every other selector in the column (paired 95\% interval). Per split: \cref{sec:app_m2w}.}
\label{tab:m2w}
\setlength{\tabcolsep}{3.5pt}
\begin{tabular}{lccc}
\toprule
\emph{higher is better} & \multicolumn{2}{c}{Policy: GUI-R1} & Qwen2.5-VL \\
\cmidrule(lr){2-3}\cmidrule(lr){4-4}
Selector & 3B & 7B & 3B \\
\midrule
Policy alone (greedy) & 38.0 & 47.8 & 19.3 \\
Majority vote of 9 & 42.5 & 49.8 & 20.4 \\
Region vote of 9 \cite{du2026guirc} & 42.5 & \underline{50.3} & 23.1 \\
CATTS \cite{lee2026catts} & 42.3 & 48.4 & 21.1 \\
Qwen3-VL-8B judge & \underline{46.0} & 49.5 & 28.4 \\
Qwen3-VL-235B judge & 41.1 & 49.3 & 23.6 \\
GUI-Critic-R1 \cite{wanyan2025guicritic} & 39.3 & 48.5 & 26.6 \\
ICM-r2 & 45.3 & \underline{50.3} & \underline{30.8} \\
GUI-Actor verifier & 35.9 & 45.3 & 26.6 \\
\midrule
\ours \method (ours) & \textbf{50.8}\rlap{*} & \textbf{53.5}\rlap{*} & \textbf{32.5}\rlap{*} \\
\bottomrule
\end{tabular}

\end{minipage}\hfill
\begin{minipage}[c]{0.485\linewidth}
\centering
\includegraphics[width=\linewidth]{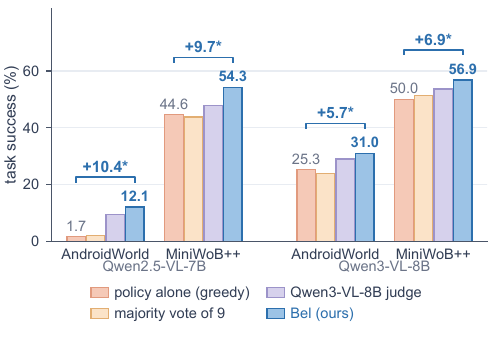}
\captionof{figure}{\textbf{Live, \method significantly raises task success in all four settings and significantly beats majority voting in each.} Task success (\%; higher is better), three runs per template; bracket: gain over greedy, star: 95\% interval excludes zero (\cref{sec:app_online}).}
\label{fig:online}
\end{minipage}
\end{figure}

\paragraph{\method significantly improves every policy.}
On offline GUI test sets, our main test, \method significantly raises the mean step success and the grounding accuracy of all three policies (\cref{tab:main}).
The gain is significant on mobile, web and desktop screens alike, for every policy (\cref{fig:hero}).
More whole tasks are right as well: on the complete AC-High test set, \method significantly raises the share of GUI-R1-3B's tasks with every step right (\cref{sec:app_types}).
On average over the six sets, the gain is largest for the general-purpose Qwen2.5-VL-3B, whose greedy action is wrong most often, and stays largest on the steps where that answer is well formed (\cref{sec:app_perset}).
For the much larger Qwen3-VL-235B-A22B, \method significantly raises step success on each of the six sets and on Multimodal-Mind2Web (\cref{sec:app_q235}), and robot-pointing success pooled over five sets (\cref{fig:embodied}b).

\paragraph{\method beats voting, CATTS and three judges.}
For every policy of \cref{tab:main}, given the same candidates (and, for the judges, the same override rule), \method has a higher mean step success and a higher grounding accuracy on both ScreenSpot sets than majority voting, region voting, CATTS, GUI-Critic-R1 and both prompted judges, significantly in every case (\cref{tab:main}, \cref{fig:hero}).
On Multimodal-Mind2Web \cite{deng2023mind2web,zheng2024seeact}, with unchanged thresholds, \method has a significantly higher step success (element, operation and value right) than every other selector of \cref{tab:m2w} for every policy, including ICM-r2 and GUI-Actor's click verifier (per split: \cref{sec:app_m2w}).

\subsection{Live Evaluation}
\label{sec:online}

\paragraph{Task success rises significantly in every setting.}
Live, in a closed loop, \method significantly raises task success over greedy decoding for both policies on both benchmarks (\cref{fig:online}).
It also significantly beats majority voting in every setting, and the Qwen3-VL-8B judge on MiniWoB++ for Qwen2.5-VL-7B, although its thresholds were chosen offline for another policy.

\paragraph{Agents act live in Android emulators.}
We evaluate on AndroidWorld \cite{rawles2025androidworld} (116 tasks in 20 apps) and on its port of MiniWoB++ \cite{liu2018miniwob} (92 tasks); each task's own test checks success.
Qwen2.5-VL-7B-Instruct and Qwen3-VL-8B-Instruct act in their own action formats; each step chooses among the greedy action and, as in GUI-PRA \cite{xiong2025guipra}, eight samples at temperature 1.0; \method and the Qwen3-VL-8B judge keep GUI-R1-3B's thresholds (\cref{sec:app_online}).

\subsection{Another Visual Agent: Robot Pointing}
\label{sec:embodied}

\begin{figure}[t]
\centering
\includegraphics[width=0.6\linewidth]{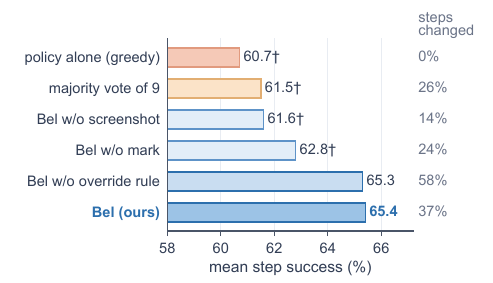}
\caption{\textbf{Drawing the action drives the gain, and the override rule keeps the best mean while changing far fewer steps.} GUI-R1-3B, mean step success over the six test sets (higher is better); each \method row removes one component; \dag: significantly below \method; right: share of steps not greedy's.}
\label{fig:ablation}
\end{figure}
\begin{figure}[t]
\centering
\includegraphics[width=\linewidth]{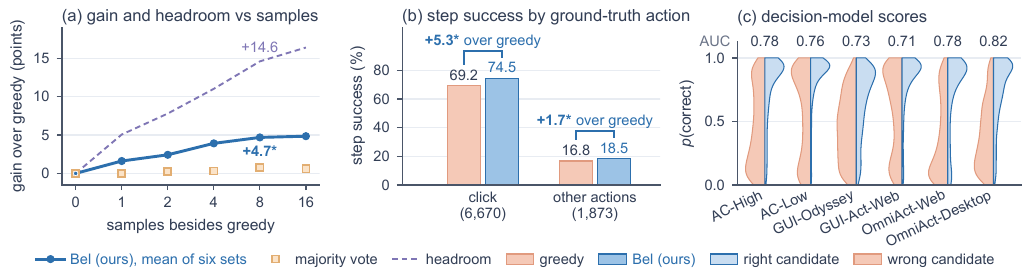}
\caption{\textbf{More samples help \method more than voting, most of the gain is on clicks, and a right candidate usually scores higher than a wrong one.} GUI-R1-3B, six sets; labels: gain over greedy (star: 95\% interval excludes zero); higher is better. (a) Gain vs samples; dashed: headroom (gain if a candidate is right). (b) By ground-truth action, steps pooled. (c) $p(\text{correct})$ of right and wrong candidates, AUC above.}
\label{fig:analysis}
\end{figure}

\paragraph{\method transfers from screens to robot pointing.}
A model that points on a robot's camera image to show it where to pick up or place an object is a visual agent of another kind.
Like a tap, its point is right or wrong by what lies under it, which the drawn mark shows to the decision model (\cref{fig:embodied}a).
Pooled over five robot-pointing sets, \method picks significantly better points than the Qwen3-VL-8B judge for all four policies, and than greedy decoding and self-consistency (the medoid point) for three of them (\cref{fig:embodied}b, \cref{sec:app_embodied}).
Pooled over two more benchmarks, Part-Afford \cite{yuan2026embodiedr1} and Point-It-Out \cite{xue2025pointitout}, it is significantly ahead of greedy decoding, majority vote, self-consistency and the judge for each of the three smaller policies (\cref{sec:app_embodied}).

\paragraph{Robot pointing tests the same principle beyond screens.}
Each candidate point is marked and asked about with the same questions as a tap; the decision model's image label and the judge's prompt name the camera view, and GUI-Critic-R1, trained on GUI steps only, is not used.
We evaluate on five robot-pointing sets from four benchmarks: the location and placement splits of RefSpatial-Bench \cite{zhou2025roborefer}, Where2Place \cite{yuan2024robopoint}, the context split of RoboSpatial-Home \cite{song2025robospatial} and VABench-Point \cite{yuan2026fsd}; a point is right when it lies in the target region.
Four policies, Qwen2.5-VL-3B, Qwen2.5-VL-7B, Qwen3-VL-8B and Qwen3-VL-235B-A22B, give the greedy answer and eight samples per item, and an answer's first point is the action; \method keeps GUI-R1-3B's AC-High thresholds (\cref{sec:app_embodied}).
Each policy gets the prompt with which RoboRefer \cite{zhou2025roborefer} evaluates Qwen models, and the points are merged on the 1\% grid, as clicks are; majority vote counts these cells.

\subsection{Ablation Study}
\label{sec:ablation}

\paragraph{Drawing the action drives the gain.}
Averaged over the six test sets, when the decision model reads the action only as text, it regains only a little of the gain; with the screenshot but the click given only as coordinates, it recovers more, and only drawing the action on the screen gives the full gain, significantly above both reduced inputs (\cref{fig:ablation}).
For actions with a point, the decision model must see what lies under the mark, not only its coordinates.
How the mark itself is drawn matters little: half or twice its size, its outline alone or a green circle each keep a significant gain over greedy decoding (GUI-R1-3B, \cref{sec:app_robust}).

\paragraph{The override rule keeps the gain of the highest score while changing far fewer steps.}
With the override rule, \method attains the highest mean step success of every variant while changing far fewer steps than taking the highest-scored candidate directly (\cref{fig:ablation}, \cref{sec:app_ablation}).
Thresholds cross-fitted on each held-out set never beat the AC-High ones (GUI-R1-3B, \cref{sec:app_ablation}).
Nor does the gain hinge on the chosen pair: every pair of the threshold grid beats greedy decoding for each policy of \cref{tab:main} (\cref{sec:app_robust}).

\subsection{Analysis and Case Study}
\label{sec:quant}
\label{sec:cases}
\label{sec:qual}

\begin{figure}[t]
\centering
\includegraphics[width=\linewidth]{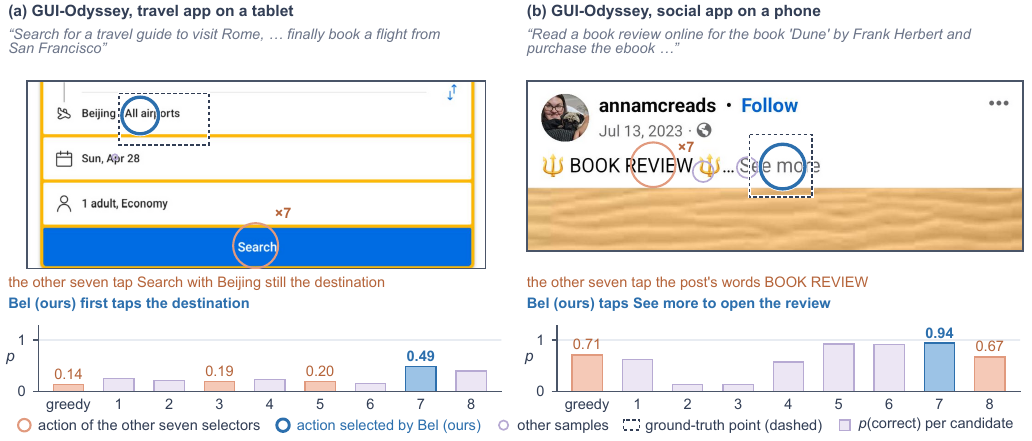}
\caption{\textbf{Two steps where every other selector of \cref{tab:main} errs and \method is right.} GUI-R1-3B; orange rings: the other selectors' actions; blue: \method's; dashed box: ground truth; bars: each candidate's $p(\text{correct})$ (rule: \cref{sec:app_cases}).}
\label{fig:cases}
\end{figure}
\begin{figure}[t]
\begin{minipage}[c]{0.485\linewidth}
\centering
\includegraphics[width=\linewidth]{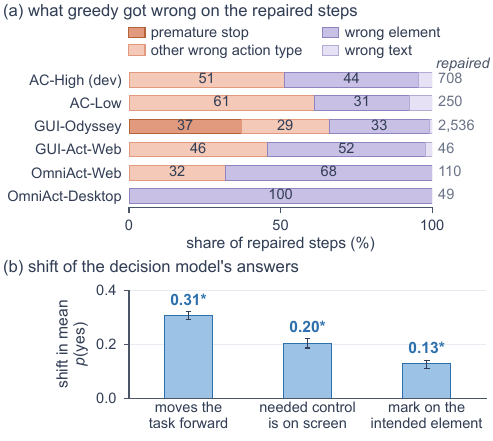}
\captionof{figure}{\textbf{\method repairs wrong action types and wrong elements.} GUI-R1-3B, complete test sets. (a) What greedy got wrong on repaired steps. (b) Shift of the decision model's answers from greedy to the selected action (95\% intervals; higher is better).}
\label{fig:qual}
\end{minipage}\hfill
\begin{minipage}[c]{0.485\linewidth}
\centering
\includegraphics[width=\linewidth]{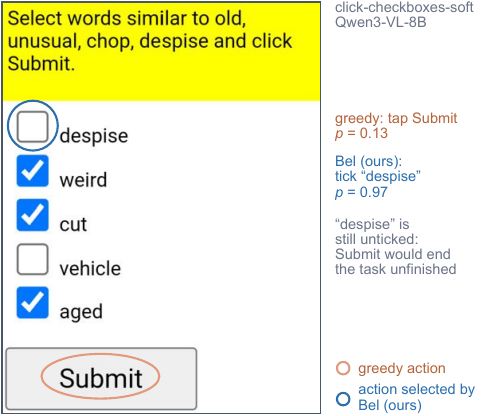}
\captionof{figure}{\textbf{Live, the decision model scores a premature Submit low and selects a word the task still needs.} MiniWoB++, Qwen3-VL-8B; the episode succeeded where greedy's failed, and this is the first step \method changed, so every earlier action is greedy's own (rule: \cref{sec:app_cases}).}
\label{fig:online_case}
\end{minipage}
\end{figure}
\begin{figure}[t]
\centering
\includegraphics[width=\linewidth]{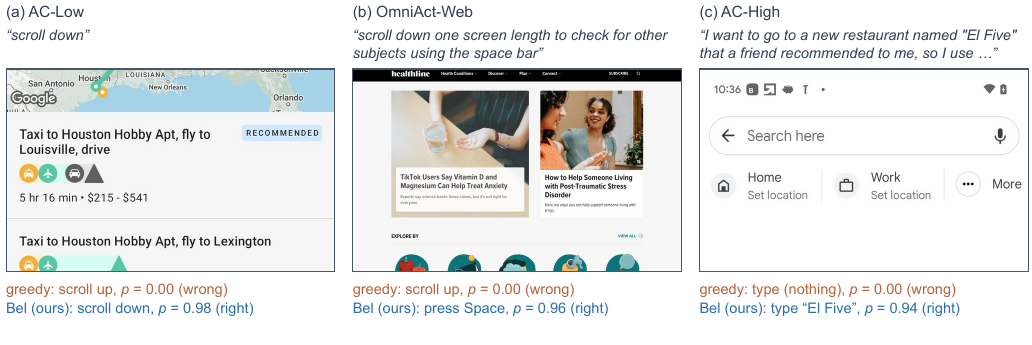}
\caption{\textbf{One probability scale also repairs actions without a position.} GUI-R1-3B: wrong greedy actions replaced by right scrolling, key-press and typing actions, for each kind the step with the largest gap in $p$ (rule and more: \cref{sec:app_gallery}).}
\label{fig:nonclick}
\end{figure}

\paragraph{More samples help \method far more than voting.}
On average over the six test sets, \method's gain rises with each doubling of the samples up to eight and holds at sixteen, while majority voting gains little (\cref{fig:analysis}a).
\method gains significantly on clicks and on other actions (\cref{fig:analysis}b).

\paragraph{Only the screen shows that the other selectors' actions are wrong.}
In \cref{fig:cases}, every other selector of \cref{tab:main} executes a wrong action and \method the right one (\cref{sec:app_cases}).
On a tablet, the others tap \emph{Search} while the destination is still Beijing, not Rome, and \method first taps the destination; on a phone, they tap a review post's words instead of its \emph{See more} link. In both, the votes also keep greedy's action, so neither repairs the step; each wrong action reads well as text, and only the screen shows that it is wrong (more: \cref{sec:app_gallery}).

\paragraph{The repaired mistakes are wrong action types and wrong elements.}
On the steps where \method replaced a wrong greedy action by a right one, greedy had mostly chosen the wrong kind of action or element, rarely wrong text (\cref{fig:qual}a); on GUI-Odyssey's long tasks the commonest mistake was a premature stop.
On these steps, the decision model's other answers, from the same call, also favour the selected action, most on moving the task forward and least on where the mark lands (\cref{fig:qual}b).

\paragraph{The same repair carries over to live episodes.}
Live, greedy taps \emph{Submit} with a requested word still unticked; the decision model scores that low, \method ticks the word first and the episode succeeds (\cref{fig:online_case}).

\paragraph{Right GUI actions score above wrong ones.}
One probability scale serves every kind of action: the same score repairs a scroll, a key press and a typed string (\cref{fig:nonclick}).
On the GUI-I split of GUI-Critic-Test \cite{wanyan2025guicritic}, the decision model judges single steps more accurately than GUI-Critic-R1, trained for this task, and than the proprietary models reported there (\cref{sec:app_scorers}).
Among GUI-R1-3B's candidates, a right one usually scores higher than a wrong one on each of the six test sets (\cref{fig:analysis}c), and on average the decision model tells them apart significantly more clearly than the Qwen3-VL-8B judge and GUI-Critic-R1 (\cref{sec:app_scorers}).
\mbox{Scoring} needs no local GPU and costs less than either API judge, a small fraction of the proprietary one (\cref{sec:app_resources}).

\section{Related Work}
\label{sec:related}

\paragraph{Visual agents.}
GUI agents read screenshots \cite{rawles2023aitw,cheng2024seeclick,hong2024cogagent}, are evaluated offline \cite{li2024androidcontrol,lu2024guiodyssey,chen2024guicourse,kapoor2024omniact,deng2023mind2web} or live \cite{rawles2025androidworld,liu2018miniwob}, and are fine-tuned \cite{cheng2024seeclick,qin2025uitars,wu2024osatlas,xu2024aguvis,gou2024uground} or trained with reinforcement learning \cite{luo2025guir1,lu2025uir1}, and models point on camera images to tell robots where to act \cite{yuan2024robopoint,zhou2025roborefer}; \method chooses among a trained agent's sampled actions, leaving its weights untouched.
Vision--language--action models also control robots directly, and the computation they spend at each step can be adapted at test time \cite{DBLP:conf/iclr/LiMSJTFMXWZ26,DBLP:journals/corr/abs-2605-29438}.

\paragraph{Judging GUI actions.}
GUI-Critic-R1 \cite{wanyan2025guicritic} is trained on labelled GUI steps to diagnose an action before it is executed, and GAIA \cite{wang2026gaia} trains a critic on agent actions from AndroidControl and GUI-Odyssey tasks to pick the best of eight sampled actions; V-Droid \cite{dai2025vdroid} trains a verifier for candidate actions extracted from the app's interface.
GUI-PRA \cite{xiong2025guipra} turns a large multimodal model into a process reward agent that synthesises verification criteria and calls visual tools to inspect the screen before it scores a step.
General language and vision--language models are also prompted to judge answers \cite{zheng2023judging} and agent behaviour \cite{pan2024autonomous}; \method instead judges each candidate where it acts, trains nothing on GUI steps and runs no agent loop: development steps only choose its thresholds and which answer is its score.

\paragraph{Choosing among candidate actions.}
PIVOT \cite{nasiriany2024pivot} draws sampled robot motions on the image and lets a vision--language model, acting as the policy, pick among them over several rounds; GUI-Actor \cite{wu2025guiactor} scores marked click candidates with a trained verifier, GTA1 \cite{yang2025gta1} lets a judge model choose among sampled proposals, and robot policies are steered at test time by re-ranking sampled actions with learned value functions or verifiers \cite{nakamoto2024vgps,kwok2025robomonkey}.
Other test-time grounding methods create new candidates, by zooming in \cite{luo2025regionfocus} or as points from overlapping sub-images that a prompted model ranks \cite{lian2025uiagile}.
Marks drawn on the image, such as a red circle \cite{shtedritski2023redcircle} or numbered regions \cite{yang2023setofmark}, steer a model to what is marked; \method marks each sampled point on its own copy.
\method scores every kind of action in one round and on one probability scale, changes no candidate, and keeps greedy unless another candidate leads it.

\paragraph{Test-time selection.}
Sampling several outputs and selecting one is a standard way to spend test-time compute \cite{yang2026languagemodelsgaintesttime}: by majority vote \cite{wang2023selfconsistency}, by an arbiter called only when the vote is contested \cite{lee2026catts}, for GUI grounding, by voting on the regions that sampled clicks cover \cite{du2026guirc}, by a verifier or reward model \cite{cobbe2021gsm8k,lightman2024verify}, or by search guided by a value estimate \cite{yao2023tot,zhou2024lats}.
\method selects in one round, with no look-ahead.

\paragraph{Policy improvement with and without training.}
Reinforcement learning improves a policy by updating its weights.
Deep agents learn by policy iteration, policy-gradient and actor--critic updates and behaviour selection \cite{DBLP:conf/icml/FanX22,DBLP:journals/corr/abs-2105-03923,DBLP:journals/corr/abs-2106-00707,DBLP:journals/corr/abs-2011-06752,DBLP:conf/iclr/FanZLH0ZWX23}, as reviewed in~\cite{DBLP:journals/corr/abs-2112-04145}; generative models are fine-tuned with rewards \cite{DBLP:conf/iclr/FanSCCLL25,DBLP:conf/nips/FanWCCL25,DBLP:journals/corr/abs-2510-18072,DBLP:journals/tmlr/WangFGNJL26}; language and audio models learn to reason, use tools and act as voice agents from outcome and process rewards \cite{DBLP:conf/iclr/FanRLPSBGLG26,zheng2026procedureaware,DBLP:journals/corr/abs-2604-02322,fan2026audiorsirecursivelyselfimprovingaudio,DBLP:journals/corr/abs-2608-26432}; and a reranker learns from the generator's feedback which evidence to select \cite{DBLP:journals/corr/abs-2602-03689}.
A model can also be improved without gradient steps: by spending more computation at test time \cite{yang2026languagemodelsgaintesttime,DBLP:journals/corr/abs-2602-06166}, or by building new weights from existing models, by fusing specialists \cite{yang2026trainingfreeonpolicydistillation} or merging checkpoints of one reinforcement-learning run \cite{yang2026doesscalingreinforcementlearning}.
\method neither trains nor builds new weights: it improves a frozen visual agent by choosing among the policy's own sampled actions at each step.

\section{Conclusion}
\label{sec:conclusion}

\method judges each action a visual agent samples where it acts: it marks each candidate on its own copy of the screen or camera image, scores every action by one probability and departs from greedy only when another candidate leads it. It needs only a frozen policy's samples and the image they act on.
For GUI agents, every policy of \cref{tab:main} gains significantly with it in step success and grounding, and significantly more than voting, CATTS and three judges on the same candidates. With GUI-R1-3B's thresholds unchanged, \method significantly raises live task success for both live policies and robot-pointing success for three policies on each of two benchmark groups.
On GUI-R1-3B, the full gain needs the drawn action, and \method mostly repairs wrong action types and elements.
\method thus gives step-level policy improvement of visual agents without training, with significant gains for GUI agents and robot pointing.

\pdfbookmark[1]{References}{references}
\bibliographystyle{unsrtcase}
\bibliography{main}

\clearpage
\appendix
\crefalias{section}{appendix}\crefalias{subsection}{subappendix}\crefalias{subsubsection}{subsubappendix}
\section{Discussion}
\label{sec:app_discussion}

\subsection{Why Judging Each Candidate Where It Acts Works}
\label{sec:app_why}

\paragraph{The mark on the screenshot carries the gain.}
Every candidate is judged where it would act: an action with a point is drawn on its own copy of the screenshot (\cref{sec:method}).
For GUI-R1-3B, averaged over the six test sets, the action given as text alone brings back only a small part of the gain, the screenshot with the click as bare coordinates brings back more, and the full gain comes only with the drawn mark, significantly above either reduced input (\cref{fig:ablation}).
The gain is insensitive to how the mark is drawn: half or twice its size, its outline alone or a green circle each keep a significant gain over greedy decoding (\cref{sec:app_robust}).

\paragraph{One absolute score, one pair of thresholds.}
Each candidate is scored alone, by the probability that it is the right next step, so taps, typing, scrolling and stopping share one scale (\cref{sec:method}).
The two thresholds are chosen per policy on AndroidControl-High and kept on every other GUI test set, Multimodal-Mind2Web included (\cref{tab:m2w}; values: \cref{sec:app_thresholds}); for robot pointing and in the live episodes, GUI-R1-3B's pair serves policies it was never chosen on (\cref{fig:embodied,fig:online}).
Every pair of the threshold grid beats greedy decoding for each policy of \cref{tab:main}, so the gain is not an artefact of the chosen pair (\cref{sec:app_robust}).

\paragraph{The greedy action changes only when another candidate leads it.}
Another candidate is executed only when the greedy action scores below $\tau$ and that candidate leads it by at least $m$ (\cref{eq:gate}). Overriding a right greedy action costs the step, and the bound $\tau$ limits this: for the three policies of \cref{tab:main}, the $\tau$ chosen on the development set is at most 0.8 (\cref{sec:app_thresholds}), so on every held-out test set only steps on which greedy scores below $\tau$ can change (on the cross-fitted AC-High, one of GUI-R1-3B's two folds selects $\tau{=}1.01$). For Qwen3-VL-235B-A22B the development set removes the bound ($\tau{=}1.01$, $m{=}0.05$; \cref{sec:app_q235}), so only the margin protects the greedy action.
For GUI-R1-3B the rule gives the highest mean step success of every variant while changing far fewer steps than always executing the highest score (\cref{fig:ablation}).

\clearpage
\subsection{Relation to Voting, Critics and Judges}
\label{sec:app_relation}

Voting counts how many samples agree and never sees the screen; trained critics and verifiers are built for the platforms and action spaces of their labels; a prompted judge returns one verdict or score per candidate.
\method instead asks a decision model for the probability that each candidate, drawn on its own copy of the screenshot as for the judges compared here, is the right next step, so one rule compares taps, typing, scrolling and stopping.
On the same candidates, \method is significantly ahead of every selector of \cref{tab:main} for each of its three policies in mean step success and on both grounding sets, including GUI-Critic-R1, which is trained partly on GUI-Odyssey.
ICM-r2 is fine-tuned on the training data of AndroidControl and GUI-Odyssey, the benchmarks of three of our six step-level test sets, so we compare with it outside that domain, where \method is significantly ahead of it on ScreenSpot and Multimodal-Mind2Web for every policy (\cref{sec:app_selectors,sec:app_m2w}).

\clearpage
\subsection{Test-Time Compute and Future Work}
\label{sec:app_compute}

\method needs nothing from the frozen policy but its greedy action and $N{=}8$ samples per step (\cref{sec:setup}), and a decision model queried through a public API as released; on the development set, labels are used to choose the thresholds and which answer serves as the score.
Scoring needs no local GPU and costs less than the API judges (\cref{tab:cost}; version and repeatability: \cref{sec:app_prompts}).
The only added cost is test-time compute: one decision-model call per distinct candidate, 4.7 per step on average (\cref{tab:cost}).
The recipe needs only that an action can be drawn on the observation; it carries over without retuning from phone, web and desktop screens to a robot's camera view (\cref{sec:app_embodied}), and the same marked questions could score the actions of other embodied agents.

\clearpage
\section{Algorithm Pseudocode}
\label{sec:app_algorithm}

\Cref{alg:bel} gives one step of \method as \cref{sec:method,sec:setup} describe it.
It takes the instruction, the history and the current screenshot, draws the candidates from the frozen policy, scores each candidate where it would act, and returns the action the agent executes.
Nothing in it is trained; the thresholds $\tau$ and $m$ are its only per-policy setting.

\begin{algorithm}[H]
\caption{\textbf{One step of \method.} The policy $\pi$ is frozen and only sampled, and nothing is trained. The circle's radius is about 2.5\% of the screenshot's longer side, $p_k$ is the decision model's probability of the verdict \emph{correct}, and every candidate is also given to it as text; samples whose action cannot be parsed are dropped, and a greedy answer that cannot be parsed stays $a^{(0)}$, without a mark. Lines 12--17 are \cref{eq:gate}.}
\label{alg:bel}
\begin{algorithmic}[1]
\Require instruction $x$, history $h_t$, screenshot $s_t$; policy $\pi$; decision model; $N$; thresholds $\tau$, $m$
\State $a^{(0)} \gets$ greedy action of $\pi(x, h_t, s_t)$ (temperature 0)
\State $a^{(1)},\dots,a^{(N)} \gets$ $N$ samples, same prompt, temperature 1.0
\State merge same-type candidates whose points share a 1\% grid cell and whose strings or directions match
\State $a^{(0)},\dots,a^{(K-1)} \gets$ the $K \le N{+}1$ distinct candidates, greedy first, then in order of first appearance
\For{$k = 0,\dots,K-1$, in parallel}
  \State $\tilde{s}_k \gets$ a copy of $s_t$
  \If{$a^{(k)}$ has a point}
    \State draw a translucent red circle at that point on $\tilde{s}_k$
  \EndIf
  \State $p_k \gets p(\text{correct} \mid x, h_t, \tilde{s}_k, a^{(k)})$
\EndFor
\State $k^\star \gets \arg\max_k p_k$ \Comment{ties: the earlier candidate}
\If{$p_0 < \tau$ \textbf{and} $p_{k^\star} - p_0 \ge m$}
  \State \Return $a^{(k^\star)}$
\Else
  \State \Return $a^{(0)}$
\EndIf
\end{algorithmic}
\end{algorithm}

\paragraph{Score sharing across policies.}
On every offline test set, a score is computed once per screen, action and grid cell (or typed string), with the mark and text of the first candidate scored there, and shared by the policies whose candidates match it, and a merged click is evaluated at its policy's first point in the cell.

\paragraph{Threshold selection.}
The pair $(\tau, m)$ with the highest step success on the evaluation subset of AndroidControl-High, the development set, is chosen over the grid $\tau \in \{0.1,\allowbreak \dots,\allowbreak 0.8,\allowbreak 1.01\}$, $m \in \{0,\allowbreak 0.05,\allowbreak 0.1,\allowbreak 0.2,\allowbreak 0.3\}$, where $\tau{=}1.01$ removes the bound on the greedy action's score; ties go to the smaller $\tau$ and then to the smaller $m$.
The pair is then carried unchanged to every other GUI test set of the policy.
The development set's own numbers are cross-fitted in two folds by task: the steps are grouped by instruction into tasks, the tasks are split in two, the pair that does best on one half is used on the other half and the reverse, so no step picks its own pair.
The values are listed in \cref{sec:app_thresholds} for the policies of \cref{tab:main} and in \cref{sec:app_q235} for Qwen3-VL-235B-A22B.

\paragraph{Transfer to new policies and platforms.}
Each offline policy is run with sampling access only (its greedy action and samples from the same prompt) and with its own $\tau$ and $m$, chosen on its candidates on the development set; this is how GUI-R1-7B, Qwen2.5-VL-3B and Qwen3-VL-235B-A22B are run (\cref{sec:setup} and \cref{sec:app_q235}).
The score, the mark, the merging grid and the six questions (\cref{sec:app_scorers}) stay fixed; only the text that comes with the image follows the setting, and on Multimodal-Mind2Web typing and selecting candidates are also marked and kept apart by cell (\cref{sec:app_m2w,sec:app_online,sec:app_embodied}).
On Multimodal-Mind2Web each policy keeps the pair chosen on AndroidControl-High (\cref{sec:app_m2w}); in the live episodes, where a different policy acts in its own action format and each candidate is rewritten in the offline test sets' action vocabulary, and for robot pointing, where the point is drawn on the camera image and the questions are the GUI ones, the pair chosen for GUI-R1-3B on AndroidControl-High is kept without change (\cref{sec:app_online,sec:app_embodied}).

\clearpage
\section{Experimental Details}
\label{sec:app_details}

This section gives the settings of every selection experiment, the prompts and the six questions put to the decision model and the prompts of every other scorer, the thresholds of the override rule chosen on the development set for each policy of \cref{tab:main} and each scorer, and the protocol of the ScreenSpot grounding experiments.

\subsection{Evaluation Protocols and Settings}
\label{sec:app_settings}

\Cref{tab:app_settings} lists, for every selection experiment, the policies, where the thresholds come from and what the bootstrap resamples.

\begin{table}[htp]
\centering\small
\caption{\textbf{Settings of the selection experiments.} Every experiment but the sample-count sweep of \cref{fig:analysis}a (\cref{sec:app_thresholds}) takes the greedy action (temperature 0) and eight samples at temperature 1.0 from the same prompt and merges them on the 1\% grid; \method scores them with \texttt{typesafe/jev-1.14}, the screenshot's longest side at most 1568 pixels. \emph{Thresholds}: where each selector's $\tau$ and $m$ come from. \emph{Unit}: what the paired 95\% bootstrap resamples; $^a$AC-Low steps, whose instructions are per step, are grouped by their AC-High task; each test set is resampled separately. $^b$Qwen3-VL-235B-A22B: the five sets of \cref{tab:app_embodied}.}
\label{tab:app_settings}
\setlength{\tabcolsep}{3pt}
\begin{tabular}{@{}>{\raggedright\arraybackslash}p{0.22\textwidth}>{\raggedright\arraybackslash}p{0.35\textwidth}>{\raggedright\arraybackslash}p{0.2\textwidth}>{\raggedright\arraybackslash}p{0.19\textwidth}@{}}
\toprule
Experiment & Policies & Thresholds & Unit \\
\midrule
Six step-level GUI test sets, evaluation subsets & GUI-R1-3B, GUI-R1-7B, Qwen2.5-VL-3B; \mbox{Qwen3-VL-235B-A22B} (\cref{sec:app_q235}) & the policy's own, chosen on AC-High (there cross-fitted) & tasks$^a$ \\
\addlinespace
ScreenSpot, ScreenSpot-Pro & as above (\mbox{Qwen3-VL-235B-A22B}: ScreenSpot) & as above & items \\
\addlinespace
Multimodal-Mind2Web, three splits & as above & as above & tasks \\
\addlinespace
Complete test sets & GUI-R1-3B & as above & tasks$^a$ \\
\addlinespace
Live AndroidWorld and MiniWoB++ & Qwen2.5-VL-7B, Qwen3-VL-8B & GUI-R1-3B's & episodes; also task templates \\
\addlinespace
Robot pointing, seven sets & Qwen2.5-VL-3B, Qwen2.5-VL-7B, Qwen3-VL-8B, \mbox{Qwen3-VL-235B-A22B}$^b$ & GUI-R1-3B's & items \\
\bottomrule
\end{tabular}
\end{table}

\begin{table}[htp]
\centering\footnotesize
\caption{\textbf{Thresholds chosen on the development set.} The override rule's $\tau$ and $m$ for each policy and each scorer shown, applied unchanged to the held-out test sets (the Qwen3-VL-235B judge's thresholds and CATTS's: \cref{sec:app_thresholds}). $\tau{=}1.01$ removes the bound on the greedy action's score, so the rule takes the highest-scored candidate whenever it beats greedy by $m$. The Claude Sonnet 4.6 judge is run for GUI-R1-3B only.}
\label{tab:app_thresholds}
\setlength{\tabcolsep}{3pt}
\begin{tabular}{lw{c}{3.4em}w{c}{3.4em}w{c}{3.4em}w{c}{3.4em}w{c}{3.4em}w{c}{3.4em}w{c}{3.4em}w{c}{3.4em}w{c}{3.4em}w{c}{3.4em}}
\toprule
 & \multicolumn{2}{c}{Qwen3-VL-8B judge} & \multicolumn{2}{c}{GUI-Critic-R1} & \multicolumn{2}{c}{ICM-r2} & \multicolumn{2}{c}{Claude Sonnet 4.6} & \multicolumn{2}{c}{\method (ours)} \\
\cmidrule(lr){2-3}\cmidrule(lr){4-5}\cmidrule(lr){6-7}\cmidrule(lr){8-9}\cmidrule(lr){10-11}
Policy & $\tau$ & $m$ & $\tau$ & $m$ & $\tau$ & $m$ & $\tau$ & $m$ & $\tau$ & $m$ \\
\midrule
GUI-R1-3B & 1.01 & 0.00 & 0.10 & 0.00 & 0.60 & 0.00 & 0.80 & 0.00 & \cellcolor{oursbg}0.80 & \cellcolor{oursbg}0.00 \\
GUI-R1-7B & 0.60 & 0.00 & 0.10 & 0.00 & 1.01 & 0.00 & -- & -- & \cellcolor{oursbg}0.60 & \cellcolor{oursbg}0.00 \\
Qwen2.5-VL-3B & 1.01 & 0.00 & 0.10 & 0.30 & 0.50 & 0.00 & -- & -- & \cellcolor{oursbg}0.80 & \cellcolor{oursbg}0.20 \\
\bottomrule
\end{tabular}
\end{table}

\clearpage
\subsection{Decision Model and Baseline Prompts}
\label{sec:app_prompts}

\paragraph{Version, use of the screen and repeatability.}
We call Jev through OpenRouter as \texttt{typesafe/\allowbreak jev-1.14}, with the marked screenshot as an image in the state; every result is that of this version.
Three checks show that its scores follow what the screen shows. First, in the live episodes every screen is rendered during the run, and \method raises task success for both policies on both benchmarks. Second, on the GUI-I split of GUI-Critic-Test, without the screenshot, the decision model falls below a rule that reads only the action type (\cref{sec:app_scorers}). Third, we pad the screenshots of the 1,372 development steps whose candidates differ in outcome (on the others no choice can change it) with a grey border of 6\% of their longer side, shifting the click points with it: the scores move by 0.06 on average (correlation 0.96), compared with 0.04 and 0.97 between two calls on the same ScreenSpot-Pro image (below), and \method's cross-fitted step success on AC-High stays 4.5 points above greedy decoding, against 4.9 on the original screenshots.
On the 10,319 ScreenSpot-Pro candidates that were scored twice, the two probabilities differ by 0.04 on average (correlation 0.97); every table uses each candidate's first score, and using the second instead moves the grounding accuracy of \method by at most half a point (0.5 for GUI-R1-7B, 0.2 for Qwen2.5-VL-3B).

\paragraph{Decision-model prompt.}
Each candidate is one call.
The state contains the user instruction, the action history as the test file gives it (the literal ``None'' on GUI-Act-Web, OmniAct and many first steps; ``(not provided)'' on the test sets without a history field: AC-Low, ScreenSpot, ScreenSpot-Pro and robot pointing), the decision written as text with its coordinates in the shown image (for example \texttt{Action: \{\textquotesingle action\textquotesingle: \textquotesingle click\textquotesingle, \textquotesingle coordinate\textquotesingle: [412, 1088]\}}), a label stating that the screenshot is taken before the action and that a click target is marked with a translucent red circle, and the screenshot (longest side at most 1568 pixels).
The questions, all answered in the same call, are:
\begin{itemize}
\item \emph{verdict} (two answers): ``Is the agent's decision for this step correct or incorrect?''; correct: ``the decision is correct and should be executed''; incorrect: ``the decision is wrong and should be revised''. Its probability of \emph{correct} is the score $p$ used by \method.
\item \emph{target on the element} (yes/no): ``The screenshot marks the action's target with a translucent red circle or arrow. Does that marked spot land on the element this action is meant to act on: \texttt{<decision>}? Judge ONLY the target position; ignore whether the action is a good idea.''
\item \emph{advances} (yes/no): ``Does this action move the agent closer to completing the user instruction \texttt{<instruction>}? Judge ONLY progress; ignore whether the target was clicked precisely.''
\item \emph{not stuck} (yes/no): ``Looking at the action history, is this action something new rather than a repeat of a step that was already tried without effect?''
\item \emph{possible here} (yes/no): ``Is this action actually possible on the screen as shown---the control it needs is present and usable in this state?''
\item \emph{holistic} (yes/no): ``User instruction: \texttt{<instruction>}. The agent is about to do: \texttt{<decision>}. Is this decision correct?''
\end{itemize}
Only the verdict enters the selection; the analysis of the repaired steps in \cref{sec:qual} uses \emph{advances}, \emph{possible here} and \emph{target on the element} (labelled ``moves the task forward'', ``needed control is on screen'' and ``mark on the intended element'' in \cref{fig:qual}b).

\paragraph{GUI-Critic-R1.}
We use the official prompt of the GUI-Critic-R1 release with the system prompt ``You are a helpful critic agent for GUI operation.'', greedy decoding and images of at most 602,112 pixels as in the release; the score is the probability of \texttt{Correct} against \texttt{Incorrect} at the first token after \texttt{<score>}.

\paragraph{Qwen3-VL-8B judge.}
The system prompt is ``You are a careful evaluator of GUI agent actions.''; the user message is ``You are judging one step of a GUI agent. User instruction: \dots\ Action history: \dots\ Decision the agent is about to execute: \dots\ The image is the screenshot before this action. If the action has a position, it is marked with a translucent red circle. Is this decision correct, i.e.\ the right next step toward completing the instruction? Answer with exactly one word: Yes or No.'' The score is $P(\text{Yes}) / (P(\text{Yes}) + P(\text{No}))$, read from the first-token log-probabilities.

\paragraph{Region vote.}
We adapt the rule of GUI-RC \cite{du2026guirc} to the greedy action and the eight samples.
Candidates are grouped by action: clicks of one action name form one group in space, and every other action is grouped by exact match.
Each click becomes a $50{\times}50$-pixel square, GUI-RC's setting; every pixel counts the squares that cover it, and a click group's vote is the highest count.
The group with the most votes wins, ties going to greedy's group, and the executed candidate is the click nearest the area-weighted centre of the largest connected area at that count.
GUI-RC itself votes over 64 samples at temperature 0.5 and outputs the consensus region, for grounding only; here it votes over the same nine candidates as every other selector.

\paragraph{CATTS arbiter.}
When the vote is contested, CATTS prompts the policy's own model once with the screenshot as the policy saw it, the instruction, the action history and the distinct candidates, each given by the policy's own answer for it (the first of its answers in that group) and the number of the policy's answers that proposed it.
The prompt asks for the candidate that best serves the instruction by the criteria of CATTS's arbiter (progress, relevance of the targeted element to the intent, no repeated action, fit with the history, feasibility on the current screen) and for a reply giving the reasoning, the number of the pick and a confidence; it is decoded greedily, with at most 384 new tokens (1,024 for Qwen3-VL-235B-A22B, served through the API with the reply started with ``Thoughts:'').
A reply without a valid pick executes the majority vote.

\paragraph{ICM-r2.}
We use the released checkpoint with the input format of its released data (GAIA-Dataset-v1.0): its system prompt verbatim, the user message ``The goal of the task (instruction): \dots\ Action (plan) history: \dots\ Current action of the agent: \dots\ Screenshot:'' followed by the image, a click written as \texttt{Tap at [x, y]} in the screenshot's pixels and marked with a hollow red circle of 24-pixel radius as in the released images, the other actions in its vocabulary (\texttt{Type}, \texttt{Swipe to}, \texttt{Open}, \texttt{Home}, \texttt{Back}, \texttt{Enter}, \texttt{Wait}, \texttt{Complete}), a scroll written as the finger's direction as in its data (converted from the content's direction that AndroidControl, GUI-Act-Web and OmniAct use), an empty history where the test file has none, and images of 256 to 3600 patches of $28{\times}28$ pixels as in its scripts.
The score is $P(\texttt{correct}) / (P(\texttt{correct}) + P(\texttt{wrong}))$ at the first token.
ICM-r2 is trained on actions from AndroidControl and GUI-Odyssey episodes; 62 of the 1,393 distinct AC-High test instructions, compared after trimming surrounding whitespace (290 steps), and 5 of the AC-Low ones occur verbatim in its released training data, and none of GUI-Odyssey's.

\paragraph{GUI-Actor's verifier.}
As in its released code, GUI-Actor-Verifier-2B sees a hollow red circle (radius 7, width 3) at the click on the original screenshot, cropped to at most $1000{\times}1000$ pixels around it, with the prompt ``Please observe the screenshot and exame [sic] whether the hollow red circle accurately placed on the intended position in the image: \dots. Answer True or False.''; the score is $P(\texttt{True}) / (P(\texttt{True}) + P(\texttt{False}))$, and a candidate without a click point scores 0.

\paragraph{Claude Sonnet 4.6 judge.}
The proprietary judge gets the same text and mark as the Qwen3-VL-8B judge, except that its last request asks for the probability, from 0 to 100, that the decision is correct, which it returns through a forced tool call with one integer field, since its API returns no token probabilities; the score is that number divided by 100.
It is Claude Sonnet 4.6 called through AWS Bedrock at temperature 0, and the screenshot is scaled to the API's working size (longest side at most 1568 pixels, at most 1.15 megapixels) before the point is marked.
It is run for GUI-R1-3B only.

\paragraph{Qwen3-VL-235B judge.}
Qwen3-VL-235B-A22B, called through AWS Bedrock at temperature 0, gets exactly the Claude judge's request: the same text, mark, scaled screenshot and forced tool call, with the score again the returned number divided by 100.
The marked screenshot is sent as PNG or as JPEG at quality 95.
A candidate scored for one policy is not scored again for another, as for every judge.
As with Claude, the API returns no token probabilities, and the judge's stated numbers are coarse: of the 107,767 candidates it scored on the GUI test sets, 56\% get exactly 95 and 82\% one of 85, 90, 95 and 100, so candidates often tie, and ties go to the earlier candidate.

\paragraph{Which candidates the added selectors score.}
ICM-r2, GUI-Actor's verifier and the Claude Sonnet 4.6 and Qwen3-VL-235B judges score the candidates of the steps whose candidates differ in outcome; on every other step every selector gets the same result, whichever candidate it picks.

\paragraph{Multimodal-Mind2Web steps.}
The policy prompt is GUI-R1's prompt for GUI-Act-Web with the actions of its list and of its answer format replaced by click, type and select (scroll and complete are dropped, and its note that some actions need a text now names typing and selecting).
Its examples are GUI-R1's click line and the typing line of GUI-R1's training prompt, \texttt{[\{\textquotesingle action\textquotesingle: enum[\textquotesingle type\textquotesingle, \textquotesingle select\textquotesingle], \textquotesingle point\textquotesingle: [123, 300], \textquotesingle input\_text\textquotesingle: \textquotesingle shanghai shopping mall\textquotesingle\}]}, with the point of the click line in place of its $[-100, -100]$; that text occurs in no instruction, history or ground truth of the three splits, so a copy of it is never right.
For the decision model a typing or selecting candidate is written as \texttt{\{\textquotesingle action\textquotesingle: \textquotesingle type\textquotesingle, \textquotesingle coordinate\textquotesingle: [x, y], \textquotesingle text\textquotesingle: \dots\}} (\texttt{select} likewise) and the label says that the target is marked; ICM-r2 gets GAIA's \texttt{Type} string for both, with its circle on the field.

\paragraph{Multimodal-Mind2Web pairs.}
Each pair shows one page twice, cropped to a 1280$\times$960 viewport at a random offset, with the same task, history and operation; a red box marks either the ground-truth element or a wrong one (a random interactive element, or the nearest element with the same tag).
Both judges score each item on its own; the prompted judge is Claude Haiku 4.5 (\texttt{claude-haiku-4.5}) asked for a likelihood from 0 to 100 with the same text and image.

\begin{table}[htp]
\centering\small
\caption{\textbf{Unlike the critics and judges run on a local GPU, the decision model needs none, and it costs less than either API judge.} Cost of scoring the candidates of 1,000 steps (4.7 per step on average); GPU time on one GPU with batched vLLM inference, model loading included; API: price of the calls (Qwen3-VL-235B and Claude: AWS Bedrock list prices).}
\label{tab:cost}
\setlength{\tabcolsep}{4pt}
\begin{tabular}{llr}
\toprule
Selector & Runs on & Per 1,000 steps \\
\midrule
Majority or region vote of 9 & CPU & negligible \\
GUI-Critic-R1 (trained, 7B) & local GPU & 39 GPU-min \\
ICM-r2 (trained, 7B) & local GPU & 63 GPU-min \\
Qwen3-VL-8B judge & local GPU & 67 GPU-min \\
Qwen3-VL-235B judge & API & \$3.79 \\
Claude Sonnet 4.6 judge & API & \$34.93 \\
\ours \method (ours) & API & \$0.83 \\
\bottomrule
\end{tabular}
\end{table}

\clearpage
\subsection{Threshold Selection for the Override Rule}
\label{sec:app_thresholds}

\Cref{tab:app_thresholds} lists, for each policy of \cref{tab:main}, the setting chosen on the development set for the Qwen3-VL-8B judge, GUI-Critic-R1, ICM-r2, the Claude Sonnet 4.6 judge and \method; those of the Qwen3-VL-235B judge and CATTS follow, and \method's own setting for Qwen3-VL-235B-A22B as the policy, $\tau{=}1.01$, $m{=}0.05$, is given with \cref{tab:app_q235} (\cref{sec:app_q235}).
For the Qwen3-VL-235B judge, the development set chooses $\tau{=}1.01$ for every policy, with $m{=}0$ for GUI-R1-3B and Qwen2.5-VL-3B and $m{=}0.05$ for GUI-R1-7B and for Qwen3-VL-235B-A22B as the policy (\cref{sec:app_q235}).
CATTS \cite{lee2026catts}, in its margin-gated form, calls its arbiter when the uncertainty of the vote, $1-\Delta$, exceeds a threshold $\tau_{\mathrm{C}}$, where $\Delta$ is the vote share of the most frequent candidate minus that of the second, and otherwise executes the majority vote; as in CATTS, the arbiter is the policy's own model, which picks one of the distinct candidates, shown with their vote counts (prompt in \cref{sec:app_prompts}), and $\tau_{\mathrm{C}}$ is chosen from CATTS's grid $0.2, 0.3, \dots, 0.8$, here on the development set.
Votes are counted over the candidates every selector shares, merged on the 1\% grid (\cref{tab:app_settings}); CATTS itself merges semantically equal actions with a language model.
The development set chooses $\tau_{\mathrm{C}}{=}0.8$ for GUI-R1-3B, GUI-R1-7B and Qwen3-VL-235B-A22B and $\tau_{\mathrm{C}}{=}0.6$ for Qwen2.5-VL-3B; at these thresholds CATTS calls its arbiter 43,300 times (on every contested step of the development set, where $\tau_{\mathrm{C}}$ is chosen, and on every other step with $1-\Delta>\tau_{\mathrm{C}}$), and on the 37 calls whose reply gives no valid pick it executes the vote.
\Cref{sec:app_selectors} adds a variant in which the Qwen3-VL-8B judge is the arbiter, executing its highest-scored candidate, and the vote is executed when $\Delta$ reaches a margin chosen from $0, \frac{1}{9}, \dots, \frac{6}{9}$ and $1.01$ (never); the development set chooses $1.01$ for both GUI-R1 policies and $\frac{6}{9}$ for Qwen2.5-VL-3B.

\Cref{fig:analysis}a uses the first $N$ samples of each step (for sixteen, the eight samples and eight more drawn with a different seed) and chooses the thresholds on AC-High again for every $N$.
A second draw of the eight samples (the second eight above), with the thresholds chosen on AC-High again, leaves \method's mean step success over the six sets at 65.4\%, as with the first draw, and its gain over greedy decoding (95\% interval 3.8 to 5.3 points) and its leads over majority and region voting stay significant.

\clearpage
\subsection{Grounding Evaluation Protocol}
\label{sec:app_screenspot}

We use the ScreenSpot and ScreenSpot-Pro test files released with GUI-R1 (1,272 and 1,581 items) and GUI-R1's ScreenSpot inference script unchanged: its prompt asks for one click, and the greedy answer is decoded as in the script; we add eight samples at temperature 1.0 from the same prompt.
ScreenSpot-Pro screenshots, up to $6016{\times}3384$ pixels, reach the policies through the script's image preprocessing, which rounds each side to a multiple of 28 pixels and downsizes the 81 screenshots above about 12.8 million pixels; as in the script, the answer's point, given in the pixels of this input, is compared with the box in the screenshot's own pixels, which lowers every selector's accuracy alike on the largest screens.
On ScreenSpot-Pro the Qwen3-VL-8B judge sees the screenshots as the decision model does, with the longest side at most 1568 pixels; on ScreenSpot, as on the step-level test sets, it is given each screenshot at its own size.
On an item where all candidates are right, or all are wrong, every selector gets the same outcome, so the baseline judges need to be run only on the other items.
An item is right when the click lies strictly inside the target element's box, as in GUI-R1's evaluation script, which reads only the point; an answer whose point can be parsed therefore counts as a click whatever it writes as the action name (Qwen2.5-VL-3B often copies the prompt's \texttt{enum['click']}).
Candidates are merged, scored and selected as on the step-level test sets, with every selector's development-set thresholds unchanged; every item is its own unit for the bootstrap.
For every policy of \cref{tab:main} and both sets, the 95\% bootstrap interval of \method's gain over greedy decoding excludes zero, and so does that of its lead over each other selector of that table: the smallest lower bounds of these leads are 2.2 (ScreenSpot) and 1.5 (ScreenSpot-Pro) points for GUI-R1-3B, 0.6 and 0.8 for GUI-R1-7B, and 7.2 and 1.6 for Qwen2.5-VL-3B. The step-success leads, with their intervals, are drawn in \cref{fig:hero}.
Weighted by the number of items, the per-category ScreenSpot-Pro accuracies that GUI-R1 reports \cite{luo2025guir1} give 28.6, 31.3 and 14.1 for GUI-R1-3B, GUI-R1-7B and Qwen2.5-VL-3B, against 28.5, 31.1 and 13.8 for our greedy runs.

\clearpage
\subsection{Computational Resources}
\label{sec:app_resources}

\method trains nothing.
The open policies, also as CATTS's arbiter, and the local judges (GUI-Critic-R1, ICM-r2, GUI-Actor's verifier and the Qwen3-VL-8B judge) run with batched inference on one GPU per model; Qwen3-VL-235B-A22B, as a policy, as CATTS's arbiter and as a judge, and the Claude judge run through AWS Bedrock.
The decision model runs through its provider's API, so \method itself needs no GPU; \cref{tab:cost} compares the scorers' costs.
CATTS uses no scorer, but calls the policy itself once more on each step or item whose vote it finds uncertain: on 434, 365 and 927 of every 1,000 steps and items of the seven held-out test sets (the five held-out step-level sets, ScreenSpot and ScreenSpot-Pro) for GUI-R1-3B, GUI-R1-7B and Qwen2.5-VL-3B.

\clearpage
\section{Additional Experimental Results}
\label{sec:app_experiments}

This section extends the experiments of the main text: what step success counts, the official script and task-level success on the complete test sets, more selectors, the three splits of Multimodal-Mind2Web, the live episodes, embodied pointing, variants of the override rule, the decision model as a judge of given actions, a much larger policy, and the robustness of the gain to how the mark is drawn and to the thresholds.
Unless stated otherwise, the settings are those of \cref{tab:app_settings}.

\subsection{What Step Success Counts}
\label{sec:app_perset}

As in GUI-R1's script, step success does not count steps whose ground truth has neither a click target nor an argument (stopping or declaring the task impossible, waiting, going back or home, key presses; 888 of GUI-R1-3B's evaluation steps); the live episodes, whose task success counts every action, measure these end to end.
Step success also ignores the action name: a point within 14\% of the screen of the ground-truth point counts as right whether the answer clicks, hovers or declares the task finished there.
If the action name must match as well, \method's gain in mean step success over the six sets is larger for every policy than in \cref{tab:main} (+5.9, +3.0 and +15.9 points for GUI-R1-3B, GUI-R1-7B and Qwen2.5-VL-3B).

\paragraph{Unparsed answers of Qwen2.5-VL-3B.}
Under GUI-R1's prompts, GUI-R1's script reads no action from 17 to 43\% of Qwen2.5-VL-3B's greedy answers per step-level test set, mostly because the model copies the prompt's \texttt{enum[...]} or writes the action in double-quoted JSON; for the two GUI-R1 policies this happens once in all (one of GUI-R1-3B's 9,431 greedy answers).
Such an answer gets no credit for its action type, but GUI-R1's script still scores its point and argument: 486 of the 2,345 such greedy answers on the scored steps of the evaluation subsets count as right.
On the steps where greedy's answer is well formed (evaluation subsets, counted as in \cref{tab:main}), the mean step success of Qwen2.5-VL-3B is 65.1 with \method and 58.5 with greedy decoding, a larger gain than for either GUI-R1 policy; over all steps, executing the first well-formed candidate whenever greedy's answer is malformed gives 50.4, against 58.8 for \method.

\clearpage
\subsection{Complete Test Sets and the Official Script}
\label{sec:app_full}

The evaluation script released with GUI-R1 \cite{luo2025guir1} drops an action category from the denominator when no step of that category is right; the main text counts every category, and the thresholds of the override rule are always chosen that way.
On the complete test sets the two ways of counting give the same step success everywhere except for \method on OmniAct-Web, and action-type accuracy differs by less than one point.

\paragraph{Task-level success.}
\label{sec:app_types}
On the complete AndroidControl-High test set, \method also raises the share of tasks with every step right for GUI-R1-3B, significantly (\cref{tab:app_types}; 95\% interval of the gain 1.5 to 3.7 points, tasks resampled).

\begin{table}[htp]
\centering\small
\caption{\textbf{\method significantly raises the share of whole tasks done right.} Share (\%) of AndroidControl-High tasks (complete test set; a task~=~all steps whose instruction is the same after trimming surrounding whitespace) in which every step is right, GUI-R1-3B: right as in step success and, on the steps that step success leaves out (neither a click target nor an argument, \cref{sec:app_perset}), with the right action type.}
\label{tab:app_types}
\begin{tabular}{lcc}
\toprule
AC-High tasks & Greedy & \method (ours) \\
\midrule
every step right (1,393 tasks) & 8.6 & \cellcolor{oursbg}11.2 \\
\bottomrule
\end{tabular}
\end{table}

\clearpage
\subsection{Comparison with Additional Selectors}
\label{sec:app_selectors}

This section compares further selectors on the candidates of \cref{tab:main} (prompts in \cref{sec:app_prompts}).
\emph{Our rule} is the override rule of \method with thresholds chosen on the development set for that selector; GAIA's rule executes the most confident candidate the critic judges correct, else greedy's; the rule of GUI-Actor's verifier takes greedy's candidate, then the samples by frequency, and executes the first whose score exceeds its threshold, else the highest-scored one.
ICM-r2 \cite{wang2026gaia} is a critic fine-tuned from Qwen2.5-VL-7B on the training data of AndroidControl and GUI-Odyssey, the benchmarks of three of the six step-level test sets, including the development set, and some of AC-High's test instructions occur in that data (\cref{sec:app_prompts}), so we compare with it on ScreenSpot and Multimodal-Mind2Web: with our rule, its ScreenSpot accuracy is 81.8, 86.1 and 71.5 for GUI-R1-3B, GUI-R1-7B and Qwen2.5-VL-3B, significantly below \method's 88.1, 89.2 and 79.4 (\cref{tab:main}). On Multimodal-Mind2Web, out of its domain, \method is significantly ahead of it for every policy (\cref{sec:app_m2w}).
GUI-Actor's verifier \cite{wu2025guiactor}, trained to judge click positions, reads a full-resolution crop around each click, and the decision model the whole screen with its longest side at most 1568 pixels; \method is significantly ahead of it on Multimodal-Mind2Web for every policy (\cref{tab:m2w}).
At about a fortieth of the cost per call (\cref{tab:cost}), \method is significantly ahead of Claude Sonnet 4.6, prompted as the Qwen3-VL-8B judge is, on GUI-Odyssey for GUI-R1-3B (step success 44.4 against 41.5).
With the Qwen3-VL-8B judge as its arbiter instead of the policy (\cref{sec:app_thresholds}), CATTS is also run on Multimodal-Mind2Web (\cref{sec:app_m2w}).

\clearpage
\subsection{Results on Multimodal-Mind2Web}
\label{sec:app_m2w}

\Cref{tab:app_m2w} adds the three test splits of Multimodal-Mind2Web \cite{deng2023mind2web,zheng2024seeact}, Cross-Task, Cross-Website and Cross-Domain, with 1,339, 1,019 and 4,060 steps of 177, 142 and 694 tasks.
No threshold is set on it: every selector keeps the thresholds chosen on AndroidControl-High (GUI-Actor's: published), and the candidates (greedy and eight samples) and their merge are those of the other test sets, except that typing and selecting candidates, like clicks, are also kept apart by their 1\% cell of the screen.

\begin{table}[htp]
\centering\small
\caption{\textbf{\method is the best selector on every Multimodal-Mind2Web split for every policy, with nothing tuned on it.} Element accuracy (the candidate's point lies inside a ground-truth element) and step success (element, operation and, for typing and selecting, the value all right), \%, averaged over the tasks of each test split (Cross-Task, Cross-Website, Cross-Domain; every step, a step without a screenshot or without its element in view counts as a failure of every selector). Every threshold is kept from AndroidControl-High for the policy, except GUI-Actor's (below); the screen is a 1280$\times$960 viewport that contains the target element when the step has one. Bold: best and underline: second best among the selectors of \cref{tab:main}; shading: gain over the policy alone (selectors of \cref{tab:main}); gain: difference of the printed values; star: its paired 95\% bootstrap interval over tasks excludes zero. The other selectors are those of \cref{sec:app_selectors}; GUI-Actor's verifier judges click positions only (ext.: our extension of it to the points of typing and selecting), and its rule uses its ScreenSpot threshold (0.8).}
\label{tab:app_m2w}
\setlength{\tabcolsep}{4pt}
\renewcommand{\arraystretch}{0.915}
\resizebox{\linewidth}{!}{%
\begin{tabular}{lccccccc}
\toprule
 & \multicolumn{2}{c}{Cross-Task} & \multicolumn{2}{c}{Cross-Website} & \multicolumn{2}{c}{Cross-Domain} & Mean \\
\cmidrule(lr){2-3}\cmidrule(lr){4-5}\cmidrule(lr){6-7}\cmidrule(lr){8-8}
Selector & Ele. Acc $\uparrow$ & Step SR $\uparrow$ & Ele. Acc $\uparrow$ & Step SR $\uparrow$ & Ele. Acc $\uparrow$ & Step SR $\uparrow$ & Step SR $\uparrow$ \\
\midrule
\multicolumn{8}{l}{\emph{Policy: GUI-R1-3B}} \\
Policy alone (greedy) & 46.4 & 38.3 & 47.8 & 36.9 & 46.3 & 38.7 & 38.0 \\
\quad Majority vote of 9 & \cellcolor[HTML]{F1F7FC}50.0 & \cellcolor[HTML]{E5EFF8}43.8 & \cellcolor[HTML]{F8FBFD}50.0 & \cellcolor[HTML]{ECF4FA}41.3 & \cellcolor[HTML]{F7FAFD}48.8 & \cellcolor[HTML]{F1F6FB}42.4 & \cellcolor[HTML]{ECF3FA}42.5 \\
\quad Region vote of 9 \cite{du2026guirc} & \cellcolor[HTML]{EEF5FB}50.6 & \cellcolor[HTML]{E5EFF8}43.8 & \cellcolor[HTML]{F9FCFE}49.8 & \cellcolor[HTML]{EFF6FB}40.8 & \cellcolor[HTML]{F7FAFD}48.8 & \cellcolor[HTML]{EEF5FB}42.9 & \cellcolor[HTML]{ECF3FA}42.5 \\
\quad CATTS \cite{lee2026catts} & \cellcolor[HTML]{ECF4FA}50.8 & \cellcolor[HTML]{E3EEF8}44.1 & \cellcolor[HTML]{EFF5FB}51.8 & \cellcolor[HTML]{EEF5FB}41.0 & \cellcolor[HTML]{FAFCFE}48.1 & \cellcolor[HTML]{F4F8FC}41.8 & \cellcolor[HTML]{EDF4FA}42.3 \\
\quad Qwen3-VL-8B judge & \cellcolor[HTML]{D0E3F3}\underline{54.5} & \cellcolor[HTML]{CBE0F2}\underline{47.0} & \cellcolor[HTML]{D5E6F4}\underline{55.4} & \cellcolor[HTML]{CFE3F3}\underline{45.1} & \cellcolor[HTML]{DEEBF7}\underline{52.7} & \cellcolor[HTML]{D9E8F5}\underline{45.8} & \cellcolor[HTML]{D1E4F4}\underline{46.0} \\
\quad Qwen3-VL-235B judge & \cellcolor[HTML]{F5F9FC}49.3 & \cellcolor[HTML]{F2F7FC}41.8 & \cellcolor[HTML]{F0F6FB}51.6 & \cellcolor[HTML]{F4F8FC}40.0 & \cellcolor[HTML]{F5F9FC}49.2 & \cellcolor[HTML]{F5F9FC}41.6 & \cellcolor[HTML]{F4F8FC}41.1 \\
\quad GUI-Critic-R1 \cite{wanyan2025guicritic} & \cellcolor[HTML]{FAFCFE}48.2 & \cellcolor[HTML]{FAFCFE}40.1 & \cellcolor[HTML]{FEFFFF}48.3 & \cellcolor[HTML]{FEFEFF}37.7 & \cellcolor[HTML]{FCFDFE}47.5 & \cellcolor[HTML]{FCFDFE}40.1 & \cellcolor[HTML]{FCFDFE}39.3 \\
\ours \quad \method (ours) & \cellcolor[HTML]{95BFE4}\textbf{60.4} & \cellcolor[HTML]{8FBCE3}\textbf{52.8} & \cellcolor[HTML]{A2C7E8}\textbf{60.6} & \cellcolor[HTML]{A4C9E8}\textbf{49.5} & \cellcolor[HTML]{B6D3ED}\textbf{57.2} & \cellcolor[HTML]{B1D0EB}\textbf{50.1} & \cellcolor[HTML]{A2C7E8}\textbf{50.8} \\
\ours \quad\emph{gain over the policy alone} & \textcolor{gain}{+14.0\rlap{*}} & \textcolor{gain}{+14.5\rlap{*}} & \textcolor{gain}{+12.8\rlap{*}} & \textcolor{gain}{+12.6\rlap{*}} & \textcolor{gain}{+10.9\rlap{*}} & \textcolor{gain}{+11.4\rlap{*}} & \textcolor{gain}{+12.8\rlap{*}} \\
\multicolumn{8}{l}{\quad\emph{other selectors, over the same candidates}} \\
\quad CATTS, Qwen3-VL-8B judge as arbiter & 54.5 & 47.0 & 55.4 & 45.1 & 52.7 & 45.8 & 46.0 \\
\quad ICM-r2, our rule & 55.9 & 47.3 & 54.9 & 43.4 & 54.2 & 45.1 & 45.3 \\
\quad ICM-r2, GAIA's rule \cite{wang2026gaia} & 53.4 & 44.8 & 53.4 & 42.8 & 52.0 & 43.1 & 43.6 \\
\quad GUI-Actor-Verifier-2B, its rule \cite{wu2025guiactor} & 42.5 & 36.3 & 41.9 & 34.4 & 43.2 & 37.0 & 35.9 \\
\quad GUI-Actor-Verifier-2B, highest score & 41.7 & 35.9 & 41.4 & 33.8 & 42.1 & 36.0 & 35.2 \\
\quad GUI-Actor-Verifier-2B, all operations (ext.), its rule & 42.1 & 35.8 & 42.6 & 32.9 & 43.4 & 36.9 & 35.2 \\
\quad GUI-Actor-Verifier-2B, all operations (ext.), highest score & 41.1 & 34.7 & 42.1 & 32.9 & 42.0 & 35.6 & 34.4 \\
\quad Oracle: any candidate right & 73.3 & 65.2 & 73.3 & 61.7 & 69.8 & 62.6 & 63.2 \\
\midrule
\multicolumn{8}{l}{\emph{Policy: GUI-R1-7B}} \\
Policy alone (greedy) & 55.8 & 48.4 & 56.4 & 46.6 & 55.9 & 48.3 & 47.8 \\
\quad Majority vote of 9 & \cellcolor[HTML]{E8F1F9}58.5 & \cellcolor[HTML]{E2EEF8}51.5 & \cellcolor[HTML]{F4F9FC}58.0 & \cellcolor[HTML]{F2F7FC}48.4 & \cellcolor[HTML]{FEFEFF}56.3 & \cellcolor[HTML]{F8FBFD}\underline{49.5} & \cellcolor[HTML]{F0F6FB}49.8 \\
\quad Region vote of 9 \cite{du2026guirc} & \cellcolor[HTML]{DEEBF7}\underline{59.2} & \cellcolor[HTML]{DBEAF6}\underline{52.0} & \cellcolor[HTML]{ECF3FA}\underline{58.8} & \cellcolor[HTML]{E5F0F9}\underline{49.5} & \cellcolor[HTML]{FDFEFF}56.4 & \cellcolor[HTML]{F8FBFD}\underline{49.5} & \cellcolor[HTML]{EAF3FA}\underline{50.3} \\
\quad CATTS \cite{lee2026catts} & \cellcolor[HTML]{F2F7FC}57.6 & \cellcolor[HTML]{F6FAFD}49.8 & \cellcolor[HTML]{F3F8FC}58.1 & 46.6 & \cellcolor[HTML]{FDFEFE}56.5 & \cellcolor[HTML]{FEFEFF}48.7 & \cellcolor[HTML]{FDFEFE}48.4 \\
\quad Qwen3-VL-8B judge & \cellcolor[HTML]{EEF5FB}58.0 & \cellcolor[HTML]{EEF5FB}50.6 & \cellcolor[HTML]{EEF5FB}58.6 & \cellcolor[HTML]{F3F8FC}48.3 & \cellcolor[HTML]{F0F6FB}\underline{57.9} & \cellcolor[HTML]{F8FBFD}\underline{49.5} & \cellcolor[HTML]{F3F8FC}49.5 \\
\quad Qwen3-VL-235B judge & \cellcolor[HTML]{F2F7FC}57.6 & \cellcolor[HTML]{EEF5FB}50.6 & \cellcolor[HTML]{F2F7FC}58.2 & \cellcolor[HTML]{F5F9FD}48.1 & \cellcolor[HTML]{F7FAFD}57.2 & \cellcolor[HTML]{FBFCFE}49.2 & \cellcolor[HTML]{F5F9FD}49.3 \\
\quad GUI-Critic-R1 \cite{wanyan2025guicritic} & \cellcolor[HTML]{F2F7FC}57.6 & \cellcolor[HTML]{F1F7FC}50.3 & \cellcolor[HTML]{FDFEFE}57.0 & \cellcolor[HTML]{FFFFFF}46.5 & \cellcolor[HTML]{FBFCFE}56.8 & \cellcolor[HTML]{FDFEFF}48.8 & \cellcolor[HTML]{FCFDFE}48.5 \\
\ours \quad \method (ours) & \cellcolor[HTML]{A2C7E8}\textbf{62.6} & \cellcolor[HTML]{AACCEA}\textbf{54.8} & \cellcolor[HTML]{8FBCE3}\textbf{64.1} & \cellcolor[HTML]{ACCDEA}\textbf{52.9} & \cellcolor[HTML]{C4DCF0}\textbf{60.9} & \cellcolor[HTML]{CDE1F2}\textbf{52.8} & \cellcolor[HTML]{B8D4ED}\textbf{53.5} \\
\ours \quad\emph{gain over the policy alone} & \textcolor{gain}{+6.8\rlap{*}} & \textcolor{gain}{+6.4\rlap{*}} & \textcolor{gain}{+7.7\rlap{*}} & \textcolor{gain}{+6.3\rlap{*}} & \textcolor{gain}{+5.0\rlap{*}} & \textcolor{gain}{+4.5\rlap{*}} & \textcolor{gain}{+5.7\rlap{*}} \\
\multicolumn{8}{l}{\quad\emph{other selectors, over the same candidates}} \\
\quad CATTS, Qwen3-VL-8B judge as arbiter & 60.7 & 52.5 & 59.5 & 49.1 & 58.4 & 49.4 & 50.3 \\
\quad ICM-r2, our rule & 59.2 & 52.1 & 58.8 & 47.9 & 58.7 & 49.8 & 50.0 \\
\quad ICM-r2, GAIA's rule \cite{wang2026gaia} & 57.9 & 50.7 & 60.0 & 50.2 & 58.0 & 49.8 & 50.3 \\
\quad GUI-Actor-Verifier-2B, its rule \cite{wu2025guiactor} & 53.5 & 47.0 & 51.6 & 42.7 & 53.5 & 46.3 & 45.3 \\
\quad GUI-Actor-Verifier-2B, highest score & 53.3 & 46.8 & 50.8 & 41.9 & 52.5 & 45.4 & 44.7 \\
\quad GUI-Actor-Verifier-2B, all operations (ext.), its rule & 54.4 & 47.4 & 53.6 & 42.5 & 53.5 & 46.3 & 45.4 \\
\quad GUI-Actor-Verifier-2B, all operations (ext.), highest score & 54.1 & 46.2 & 52.1 & 41.9 & 52.3 & 45.3 & 44.5 \\
\quad Oracle: any candidate right & 74.1 & 68.5 & 75.7 & 68.8 & 72.6 & 66.9 & 68.1 \\
\midrule
\multicolumn{8}{l}{\emph{Policy: Qwen2.5-VL-3B}} \\
Policy alone (greedy) & 34.2 & 20.1 & 34.7 & 16.7 & 34.6 & 21.1 & 19.3 \\
\quad Majority vote of 9 & \cellcolor[HTML]{FFFFFF}34.9 & \cellcolor[HTML]{FEFEFF}21.2 & \cellcolor[HTML]{FFFFFF}35.1 & \cellcolor[HTML]{FFFFFF}17.2 & \cellcolor[HTML]{FEFEFF}35.7 & \cellcolor[HTML]{FDFEFF}22.8 & \cellcolor[HTML]{FEFEFF}20.4 \\
\quad Region vote of 9 \cite{du2026guirc} & \cellcolor[HTML]{FCFDFE}36.8 & \cellcolor[HTML]{F8FBFD}24.3 & \cellcolor[HTML]{FEFFFF}35.7 & \cellcolor[HTML]{FAFCFE}20.0 & \cellcolor[HTML]{FCFDFE}36.8 & \cellcolor[HTML]{F9FBFD}25.0 & \cellcolor[HTML]{F9FBFE}23.1 \\
\quad CATTS \cite{lee2026catts} & \cellcolor[HTML]{FEFEFF}35.4 & \cellcolor[HTML]{FBFDFE}22.9 & \cellcolor[HTML]{FFFFFF}34.0 & \cellcolor[HTML]{FEFFFF}17.5 & \cellcolor[HTML]{FFFFFF}34.4 & \cellcolor[HTML]{FDFEFF}22.9 & \cellcolor[HTML]{FDFEFF}21.1 \\
\quad Qwen3-VL-8B judge & \cellcolor[HTML]{F7FAFD}\underline{38.7} & \cellcolor[HTML]{E4EFF8}\underline{30.6} & \cellcolor[HTML]{FEFFFF}35.6 & \cellcolor[HTML]{ECF4FA}\underline{24.9} & \cellcolor[HTML]{FCFDFE}\underline{36.9} & \cellcolor[HTML]{EAF3FA}\underline{29.8} & \cellcolor[HTML]{E9F2F9}\underline{28.4} \\
\quad Qwen3-VL-235B judge & \cellcolor[HTML]{FFFFFF}34.7 & \cellcolor[HTML]{F7FAFD}24.9 & \cellcolor[HTML]{FEFEFF}\underline{35.9} & \cellcolor[HTML]{F7FAFD}21.4 & \cellcolor[HTML]{FFFFFF}34.2 & \cellcolor[HTML]{FAFCFE}24.6 & \cellcolor[HTML]{F8FBFD}23.6 \\
\quad GUI-Critic-R1 \cite{wanyan2025guicritic} & \cellcolor[HTML]{FDFEFE}36.1 & \cellcolor[HTML]{EFF5FB}27.6 & \cellcolor[HTML]{FFFFFF}34.4 & \cellcolor[HTML]{ECF4FA}24.8 & \cellcolor[HTML]{FFFFFF}35.1 & \cellcolor[HTML]{F2F7FC}27.4 & \cellcolor[HTML]{EFF6FB}26.6 \\
\ours \quad \method (ours) & \cellcolor[HTML]{E8F1F9}\textbf{43.5} & \cellcolor[HTML]{DAE9F6}\textbf{33.0} & \cellcolor[HTML]{E9F2FA}\textbf{43.7} & \cellcolor[HTML]{D5E6F5}\textbf{30.6} & \cellcolor[HTML]{EDF4FA}\textbf{42.6} & \cellcolor[HTML]{DAE9F6}\textbf{33.9} & \cellcolor[HTML]{D9E8F5}\textbf{32.5} \\
\ours \quad\emph{gain over the policy alone} & \textcolor{gain}{+9.3\rlap{*}} & \textcolor{gain}{+12.9\rlap{*}} & \textcolor{gain}{+9.0\rlap{*}} & \textcolor{gain}{+13.9\rlap{*}} & \textcolor{gain}{+8.0\rlap{*}} & \textcolor{gain}{+12.8\rlap{*}} & \textcolor{gain}{+13.2\rlap{*}} \\
\multicolumn{8}{l}{\quad\emph{other selectors, over the same candidates}} \\
\quad CATTS, Qwen3-VL-8B judge as arbiter & 38.7 & 30.6 & 35.6 & 24.9 & 36.9 & 29.8 & 28.4 \\
\quad ICM-r2, our rule & 39.3 & 32.6 & 37.4 & 28.3 & 38.2 & 31.4 & 30.8 \\
\quad ICM-r2, GAIA's rule \cite{wang2026gaia} & 40.7 & 28.7 & 39.3 & 25.0 & 39.7 & 28.8 & 27.5 \\
\quad GUI-Actor-Verifier-2B, its rule \cite{wu2025guiactor} & 32.7 & 27.1 & 32.3 & 24.9 & 33.1 & 27.9 & 26.6 \\
\quad GUI-Actor-Verifier-2B, highest score & 33.0 & 27.4 & 31.8 & 24.4 & 33.1 & 27.9 & 26.5 \\
\quad GUI-Actor-Verifier-2B, all operations (ext.), its rule & 33.5 & 25.6 & 34.0 & 24.0 & 32.5 & 26.3 & 25.3 \\
\quad GUI-Actor-Verifier-2B, all operations (ext.), highest score & 33.3 & 25.3 & 33.2 & 23.6 & 32.1 & 25.8 & 24.9 \\
\quad Oracle: any candidate right & 53.6 & 39.0 & 53.9 & 35.9 & 52.0 & 39.9 & 38.2 \\
\bottomrule
\end{tabular}}
\end{table}

Each step shows a 1280$\times$960 viewport of the page that contains the target element, at a vertical offset drawn at random for each step (the top of the page when the step has no target element in the data); the instruction is the task, and the history lists the previous actions in GUI-R1's format.
The policies get GUI-R1's prompt for GUI-Act-Web with the actions click, type and select (\cref{sec:app_prompts}).
On Multimodal-Mind2Web typing and selecting act on an element, so a typing or selecting candidate is marked at its point and its decision string carries the point, for the decision model and the prompted judges alike; on the other test sets such a candidate has no mark.
The element is right when the candidate's point lies inside a ground-truth element (a candidate without a point never is), and the step is right when, in addition, the operation is right and, for typing and selecting, the value equals the ground truth (Multimodal-Mind2Web's value or the value as written in its action text) after lower-casing and trimming.
Each metric is averaged over the steps of a task and then over the tasks.
Every step counts: the 11 steps without a screenshot and the 352 without a ground-truth element in view are failures of every selector.
The scored baselines judge the candidates of the steps whose candidates differ in outcome; on the other steps every selector gets the same result.

\method raises the step success of every policy of \cref{tab:app_m2w} significantly on every split (95\% paired bootstrap over tasks), and for each of them it has the highest step success of all the selectors but the oracle on every split and on average.
On average it is significantly ahead of every other selector for every policy.
ICM-r2, trained in domain for three of the six step-level test sets (\cref{sec:app_selectors}), is significantly behind \method on Multimodal-Mind2Web, out of its domain, for every policy.
GUI-Actor's verifier, which judges click positions only, lowers the step success of both GUI-R1 policies, and extending it to the points of typing and selecting does not help.
No parsed greedy action types the text of the prompt's typing example; parsed samples of Qwen2.5-VL-3B do on 26 steps in all, and such a copy is never right.

\clearpage
\subsection{Live Evaluation}
\label{sec:app_online}

\paragraph{Agent harness.}
Both policies act in their own mobile action format, taken from the agent examples released with the models.
Qwen2.5-VL-7B receives the \texttt{mobile\_use} tool with the resolution of the screenshot it sees (the screen at the processor's default resolution, with coordinates in its pixels) and the list of its earlier tool calls.
We add the \texttt{answer} action of the later version of the same tool, because AndroidWorld scores information tasks on the reported answer, and map the few action names the model borrows from other tools (\texttt{left\_click}, \texttt{tap}, \texttt{long\_click}, \texttt{input\_text}, \texttt{open\_app}) to their \texttt{mobile\_use} equivalents, the same for every selector.
Qwen3-VL-8B receives the system prompt of its example unchanged, with coordinates on a 0--999 grid, and its own one-line descriptions of the earlier actions.
On MiniWoB++ the policy, but not the scorers, is also given AndroidWorld's guideline to stay in the task app.
Actions are executed through AndroidWorld's device interface, and the step budget is AndroidWorld's, ten steps per unit of task complexity.
An AndroidWorld episode can end in an infrastructure error, almost always before the agent acts: the task's setup fails, or the emulator stops responding or loses its accessibility channel, which happens once Wi-Fi is off.
Such an episode says nothing about the agent; it is rerun once with the same task instance in a fresh emulator, for every selector alike, with every Wi-Fi task in an emulator of its own, and counts as a failure if it errs again.
A MiniWoB++ episode whose reward is not an integer, or whose reward can no longer be read because the agent left the task app, is a failure, since AndroidWorld's MiniWoB++ tasks count only a reward of exactly one as success.
A composite AndroidWorld task scores the mean of its subtasks and, as in AndroidWorld's verdict on each episode, succeeds only when this mean exceeds one half.
Every task template is run for each policy with three seeds, AndroidWorld's default task seed and two others, and the results are pooled; the third seed, used alike for both policies, both benchmarks and every selector, was added after the first two had been analysed.
The seed fixes the parameters of AndroidWorld's tasks, which every selector shares; a MiniWoB++ page draws a new instance of its template whenever it starts, so MiniWoB++ episodes are paired by template.

\paragraph{Candidate scoring.}
As in the offline experiments, candidates are merged on a 1\% grid of the screen; a swipe is merged on the cells of its start and end points.
A step whose greedy output cannot be parsed does nothing, and no samples are drawn for it; samples that cannot be parsed are dropped.
The decision model receives the instruction, the earlier actions as AndroidWorld actions (the action type with its typed text, app name or scroll direction, but not where a click landed; offline, the history is what GUI-R1's test files give: plain-language step descriptions on AndroidControl-High, action strings on GUI-Odyssey, and none on the other sets), the screenshot (longest side at most 1568 pixels) with the candidate's point marked, and the candidate in the action vocabulary of the offline test sets: a swipe becomes a scroll in the opposite direction that keeps its start point, a system button becomes the matching press action (\texttt{press\_back}, \texttt{press\_home}, \texttt{press\_enter}), and ending the task keeps AndroidWorld's form, a \texttt{status} action whose goal status is complete or infeasible.
The Qwen3-VL-8B judge receives the same image, history and decision.
The thresholds are those chosen offline for GUI-R1-3B on AC-High ($\tau{=}0.8$, $m{=}0$ for the decision model; $\tau{=}1.01$, $m{=}0$ for the Qwen3-VL-8B judge).

\paragraph{Selector behaviour.}
\Cref{tab:app_online} shows that \method changes the action on a sizeable share of the live steps, and that, when another candidate is proposed, both scorers overrule most of the policy's decisions to end the task; for Qwen3-VL-8B on AndroidWorld most stops are proposed by every sample and never reach the scorer.
\method is ahead of the Qwen3-VL-8B judge in all four settings (\cref{fig:online}), significantly for Qwen2.5-VL-7B on MiniWoB++ (by 6.5 points; 95\% interval over episodes $[1.1, 12.0]$), and in every setting it changes fewer of the policy's actions than the judge (changed steps in \cref{tab:app_online}).
\method's lead over majority vote has the 95\% interval over episodes [+6.6, +13.5] (AndroidWorld) and [+5.4, +15.6] (MiniWoB++) for Qwen2.5-VL-7B, and [+2.9, +11.5] and [+0.7, +10.1] for Qwen3-VL-8B.
The intervals of \cref{fig:online} resample episodes; resampling the task templates instead, each with its three seeds, keeps significant every gain over greedy decoding and over majority vote, and the lead over the Qwen3-VL-8B judge for Qwen2.5-VL-7B on MiniWoB++ (95\% interval $[0.4, 12.7]$).
\Cref{fig:online_case} shows the step chosen by the rule of \cref{sec:app_cases}.

\begin{table}[htp]
\centering\footnotesize
\caption{\textbf{In every live setting, \method overrules most of greedy's stops that have an alternative; majority vote, at most a quarter.} \emph{Steps per episode}: over the episodes that ended without an environment error; \emph{changed steps}: share of all steps on which the executed action is not the greedy one; \emph{greedy's stops}: steps on which greedy's action ends the task, and in brackets those on which another candidate was proposed; \emph{stops overruled}: share of the latter on which the selector chose an action that does not end the task. Scoring cost: the price of the decision model's calls; the Qwen3-VL-8B judge runs on a local GPU.}
\label{tab:app_online}
\setlength{\tabcolsep}{3.4pt}
\begin{tabular}{lccccccc}
\toprule
Selector & Episodes & \shortstack{\vphantom{dy}Steps per\\\vphantom{dy}episode} & \shortstack{\vphantom{dy}Steps with\\\vphantom{dy}several\\\vphantom{dy}candidates} & \shortstack{\vphantom{dy}Changed\\\vphantom{dy}steps} & \shortstack{\vphantom{dy}Greedy's stops\\\vphantom{dy}(with an\\\vphantom{dy}alternative)} & \shortstack{\vphantom{dy}Stops\\\vphantom{dy}overruled} & \shortstack{\vphantom{dy}Scoring cost\\\vphantom{dy}per episode} \\
\midrule
\multicolumn{8}{l}{\emph{Qwen2.5-VL-7B, AndroidWorld}} \\
\quad Policy alone (greedy) & 348 & 17.2 & -- & -- & -- & -- & -- \\
\quad Majority vote of 9 & 348 & 16.4 & 85\% & 8\% & 58 (51) & 0\% & -- \\
\quad Qwen3-VL-8B judge & 348 & 12.7 & 94\% & 57\% & 449 (381) & 86\% & local GPU \\
\ours \quad \method (ours) & 348 & 13.3 & 92\% & 51\% & 521 (448) & 88\% & \$0.012 \\
\midrule
\multicolumn{8}{l}{\emph{Qwen2.5-VL-7B, MiniWoB++}} \\
\quad Policy alone (greedy) & 276 & 8.7 & -- & -- & -- & -- & -- \\
\quad Majority vote of 9 & 276 & 8.7 & 81\% & 7\% & 11 (10) & 0\% & -- \\
\quad Qwen3-VL-8B judge & 276 & 8.0 & 93\% & 44\% & 76 (70) & 89\% & local GPU \\
\ours \quad \method (ours) & 276 & 6.7 & 91\% & 39\% & 81 (71) & 87\% & \$0.005 \\
\midrule
\multicolumn{8}{l}{\emph{Qwen3-VL-8B, AndroidWorld}} \\
\quad Policy alone (greedy) & 348 & 14.5 & -- & -- & -- & -- & -- \\
\quad Majority vote of 9 & 348 & 14.3 & 50\% & 9\% & 183 (68) & 9\% & -- \\
\quad Qwen3-VL-8B judge & 348 & 14.3 & 51\% & 29\% & 236 (73) & 68\% & local GPU \\
\ours \quad \method (ours) & 348 & 14.7 & 50\% & 23\% & 248 (94) & 76\% & \$0.004 \\
\midrule
\multicolumn{8}{l}{\emph{Qwen3-VL-8B, MiniWoB++}} \\
\quad Policy alone (greedy) & 276 & 5.7 & -- & -- & -- & -- & -- \\
\quad Majority vote of 9 & 276 & 5.6 & 61\% & 10\% & 37 (16) & 25\% & -- \\
\quad Qwen3-VL-8B judge & 276 & 5.9 & 68\% & 42\% & 81 (64) & 92\% & local GPU \\
\ours \quad \method (ours) & 276 & 5.9 & 60\% & 27\% & 65 (48) & 92\% & \$0.002 \\
\bottomrule
\end{tabular}
\end{table}

\paragraph{One harness for every selector.}
Offline every selector scores the same candidates; in live episodes each selector's arm draws its own candidates in the same way, the greedy output and eight samples at temperature 1.0, and follows its own trajectory.

\begin{figure}[htp]
\centering
\includegraphics[width=\linewidth]{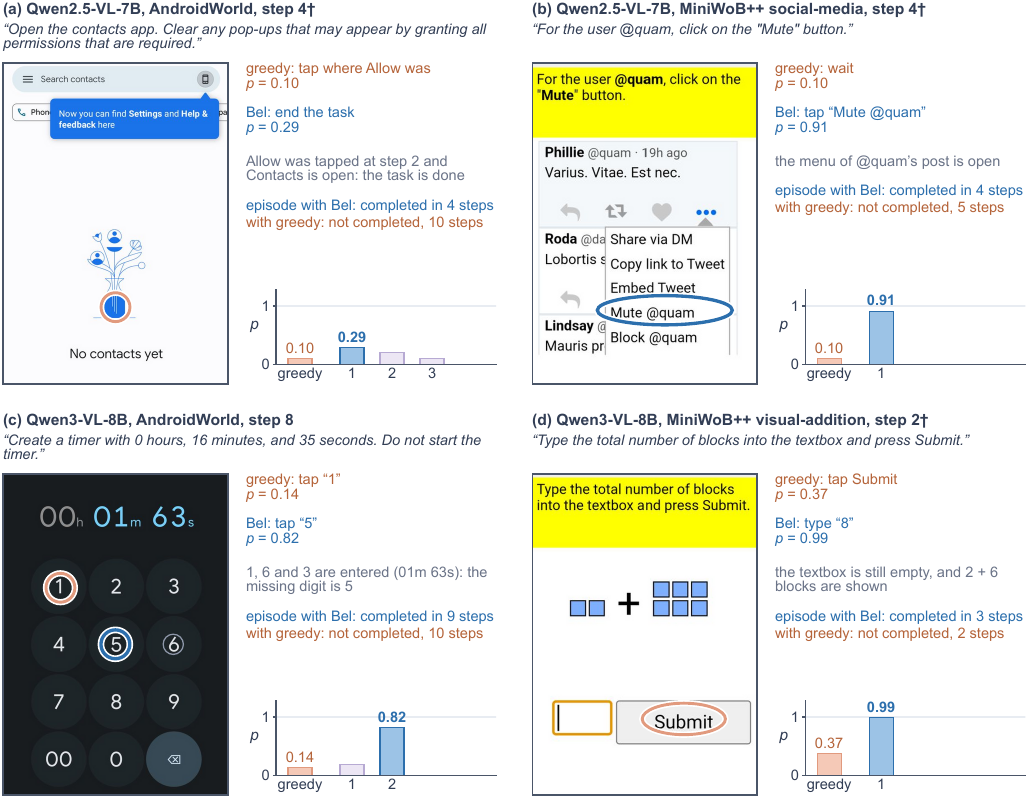}
\caption{\textbf{At its first change in these live episodes, \method replaces an idle, wrong or premature greedy action with the one the screen calls for.} One episode per policy and benchmark, shown at the first step on which \method changed the action, so every earlier action is greedy's own; beside each screen, the two actions with the decision model's $p$, our reading of the screen, and the outcome of the \method episode and of the greedy episode with the same task and task seed (on MiniWoB++, a new page of the same template). Selection, within each setting: among the episodes that \method completes and greedy decoding does not, the first changed step is taken when both actions are clicks, greedy's $p$ is below 0.3 and the selected action's above 0.8, and, by our reading of the screen, greedy's action does not also advance the task and the selected one does; the largest gap in $p$ is shown, and the step of \cref{fig:online_case} is left out. Panel (c) meets this rule; a setting without such a step ($\dagger$) relaxes it in turn: other action types are allowed (b), then both thresholds become 0.5 (d), then they are dropped (a). On AndroidWorld, a greedy upward swipe on the home screen opens the app list, from which the app opens too, so such first steps are left out. Orange ring: greedy's action; blue ring: the action \method executes; grey ring: another click candidate; waiting, typing and ending the task carry no position and no ring; bars: every distinct candidate's $p(\text{correct})$ in sampling order.}
\label{fig:gal_online}
\end{figure}

\Cref{fig:gal_online} shows one live episode for each policy and benchmark, chosen within each of the four settings by the rule of the live case (\cref{sec:app_cases}) with one added condition, stated in the caption, relaxed where a setting has no step that meets it.

\clearpage
\subsection{Transfer to Robot Pointing}
\label{sec:app_embodied}

\begin{table}[htp]
\centering\small
\caption{\textbf{Without retuning, \method is the best robot-point selector for every policy.} Success (\%) of the chosen point, pooled over the five embodied pointing sets; items per set RefSp.-Loc. 100, RefSp.-Place. 100, Where2Place 100, RoboSpatial 122, VABench-P 300 (RefSp.: RefSpatial-Bench, location and placement splits; RoboSpatial: the context split of RoboSpatial-Home; VABench-P: VABench-Point); Qwen3-VL-235B-A22B: \cref{sec:app_q235}. Majority vote over the 1\% cells; self-consistency: the medoid of the nine answers. Bold: best; underline: second best (per column, selectors only); gain: difference of the printed values; star: its paired 95\% bootstrap interval over items excludes zero.}
\label{tab:app_embodied}
\def\embpooledbody{%
\toprule
 & \multicolumn{4}{c}{Policy} \\
\cmidrule(lr){2-5}
Selector & Qwen2.5-VL-3B & Qwen2.5-VL-7B & Qwen3-VL-8B & Qwen3-VL-235B-A22B \\
\midrule
Policy alone (greedy) & 21.3 & \underline{26.9} & \underline{48.6} & \underline{56.1} \\
\quad Majority vote of 9 & 21.5 & 26.6 & 47.2 & 56.0 \\
\quad Self-consistency & 19.4 & 24.5 & 47.9 & 55.7 \\
\quad Qwen3-VL-8B judge & \underline{23.3} & 25.2 & 45.8 & 52.2 \\
\ours \quad \method (ours) & \textbf{29.1} & \textbf{35.9} & \textbf{50.8} & \textbf{60.4} \\
\ours \gainrow{gain over the policy alone} & {\footnotesize\textcolor{gain}{+7.8\rlap{*}}} & {\footnotesize\textcolor{gain}{+9.0\rlap{*}}} & {\footnotesize\textcolor{gain}{+2.2}} & {\footnotesize\textcolor{gain}{+4.3\rlap{*}}} \\
\midrule
Oracle (any candidate) & 50.7 & 58.0 & 70.2 & 76.5 \\
\bottomrule
}%
\def\embpooledpad{0pt}%
\setlength{\tabcolsep}{4pt}%
\sbox0{\begin{tabular}{l*{4}{>{\hspace{\embpooledpad}}c<{\hspace{\embpooledpad}}}}\embpooledbody\end{tabular}}%
\edef\embpooledpad{\the\dimexpr(\textwidth-\wd0)/8\relax}%
\begin{tabular}{l*{4}{>{\hspace{\embpooledpad}}c<{\hspace{\embpooledpad}}}}\embpooledbody\end{tabular}
\end{table}

\begin{table}[htp]
\centering\footnotesize
\caption{\textbf{On two more pointing benchmarks, \method significantly lifts three policies and is the best selector throughout.} Success (\%) of the chosen point; items Part-Afford 2,000, Point-It-Out 346; candidates, selectors, thresholds and markings as for the five sets of \cref{tab:app_embodied}.}
\label{tab:app_embodied_new}
\def\embnewbody{%
\toprule
Policy & \multicolumn{3}{c}{Qwen2.5-VL-3B} & \multicolumn{3}{c}{Qwen2.5-VL-7B} & \multicolumn{3}{c}{Qwen3-VL-8B} \\
\cmidrule(lr){2-4}\cmidrule(lr){5-7}\cmidrule(lr){8-10}
Selector & \shortstack{\vphantom{dy}Part-\\\vphantom{dy}Afford} & \shortstack{\vphantom{dy}Point-It-\\\vphantom{dy}Out} & Pooled & \shortstack{\vphantom{dy}Part-\\\vphantom{dy}Afford} & \shortstack{\vphantom{dy}Point-It-\\\vphantom{dy}Out} & Pooled & \shortstack{\vphantom{dy}Part-\\\vphantom{dy}Afford} & \shortstack{\vphantom{dy}Point-It-\\\vphantom{dy}Out} & Pooled \\
\midrule
Policy alone (greedy) & 25.7 & 45.7 & 28.6 & 44.8 & 52.0 & 45.8 & \underline{59.3} & 59.5 & \underline{59.3} \\
\quad Majority vote of 9 & 26.1 & 45.4 & 28.9 & \underline{45.3} & 52.3 & \underline{46.3} & 57.3 & 58.1 & 57.4 \\
\quad Self-consistency & 24.7 & 44.2 & 27.6 & 39.8 & \underline{52.6} & 41.6 & 56.8 & \underline{61.3} & 57.5 \\
\quad Qwen3-VL-8B judge & \underline{31.3} & \underline{46.8} & \underline{33.6} & 41.9 & 48.0 & 42.8 & 52.1 & 59.0 & 53.1 \\
\ours \quad \method (ours) & \textbf{44.8} & \textbf{55.5} & \textbf{46.4} & \textbf{59.0} & \textbf{59.0} & \textbf{59.0} & \textbf{62.3} & \textbf{64.5} & \textbf{62.6} \\
\ours \gainrow{gain over the policy alone} & {\scriptsize\textcolor{gain}{+19.1\rlap{*}}} & {\scriptsize\textcolor{gain}{+9.8\rlap{*}}} & {\scriptsize\textcolor{gain}{+17.8\rlap{*}}} & {\scriptsize\textcolor{gain}{+14.2\rlap{*}}} & {\scriptsize\textcolor{gain}{+7.0\rlap{*}}} & {\scriptsize\textcolor{gain}{+13.2\rlap{*}}} & {\scriptsize\textcolor{gain}{+3.0\rlap{*}}} & {\scriptsize\textcolor{gain}{+5.0\rlap{*}}} & {\scriptsize\textcolor{gain}{+3.3\rlap{*}}} \\
\midrule
Oracle (any candidate) & 64.9 & 73.1 & 66.1 & 79.6 & 77.2 & 79.2 & 84.6 & 80.3 & 84.0 \\
\bottomrule
}%
\def\embnewpad{0pt}%
\setlength{\tabcolsep}{3pt}%
\sbox0{\begin{tabular}{l*{9}{>{\hspace{\embnewpad}}c<{\hspace{\embnewpad}}}}\embnewbody\end{tabular}}%
\edef\embnewpad{\the\dimexpr(\textwidth-\wd0)/18\relax}%
\begin{tabular}{l*{9}{>{\hspace{\embnewpad}}c<{\hspace{\embnewpad}}}}\embnewbody\end{tabular}
\end{table}

\paragraph{Benchmarks.}
RefSpatial-Bench \cite{zhou2025roborefer} (location and placement splits, 100 items each; its unseen split is meant only for models trained on RefSpatial), Where2Place \cite{yuan2024robopoint} (100), the context split of RoboSpatial-Home \cite{song2025robospatial} (122) and VABench-Point \cite{yuan2026fsd} (300), from their official releases.
Each item has a camera image, an instruction and a mask of the target region.

\paragraph{Policies and candidates.}
The instruction is the benchmark's task sentence without its answer-format request, followed by ``Output the point coordinates in JSON format.'', the output request of the prompt with which RoboRefer evaluates Qwen models; the first point of an answer is the action.
Qwen2.5-VL answers in pixels of its resized input and Qwen3-VL on a 0--1000 grid; both are mapped to the original image.
Each item has the greedy point and eight samples at temperature 1.0, merged on the 1\% grid.
A point is right when it lies on the target mask, and each item is scored on one point from our own prompt (the scoring scripts of RefSpatial-Bench, Where2Place and VABench-Point average over several points per answer, and RoboSpatial-Home's accepts an answer when either of its first two points lies on the mask).

\paragraph{Selectors.}
Majority vote over the 1\% cells stays close to greedy decoding for every policy (pooled success 21.5, 26.6, 47.2 and 56.0\% for Qwen2.5-VL-3B, Qwen2.5-VL-7B, Qwen3-VL-8B and Qwen3-VL-235B-A22B, against 21.3, 26.9, 48.6 and 56.1\%), since the points of an item often fall in different cells; \emph{self-consistency} instead takes the medoid, the parsed point (greedy or sampled) with the smallest total distance to the others.
The decision model and the Qwen3-VL-8B judge see the camera image with the point marked and the instruction; the decision model's image label and the judge's whole prompt call the image the robot's camera view, and the decision model's questions are the GUI ones, unchanged.
\method keeps the thresholds chosen for GUI-R1-3B on AC-High and the judge its own; GUI-Critic-R1, trained on GUI steps only, is not used.
The decision model's API rejects images with a side below 280 pixels (its documentation), so the 48 VABench-Point frames of $320{\times}256$ or $224{\times}224$ pixels and the 157 Point-It-Out frames of $456{\times}256$ or $320{\times}256$ pixels are upscaled to a 384-pixel short side before the point is drawn.

\begin{figure}[p]
\centering
\includegraphics[width=\linewidth]{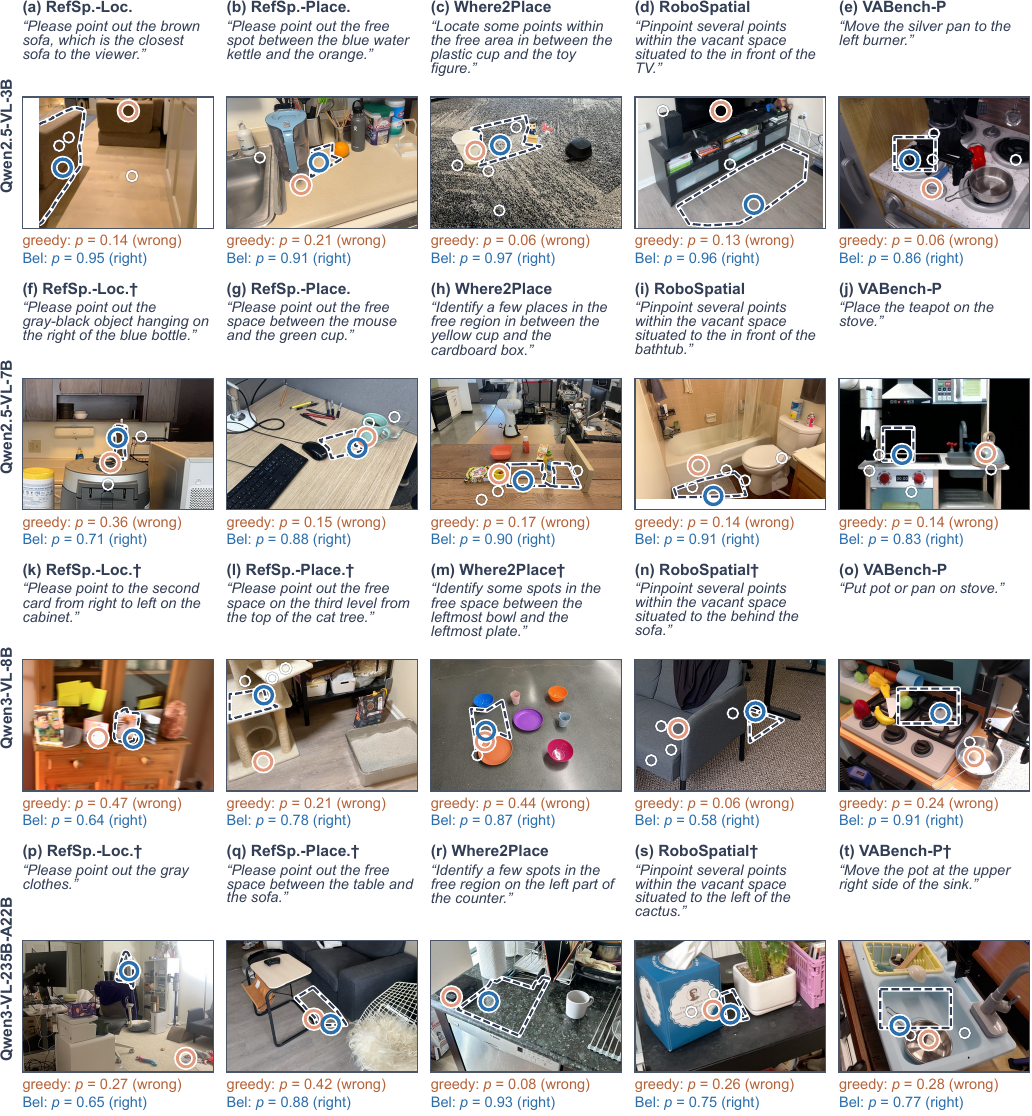}
\caption{\textbf{For every policy and every pointing set, an item on which \method moves a wrong greedy point onto the target.} One item per policy (row) and set (column), chosen by a rule: \method replaced greedy's point off the target region by a sampled point on it, $p(\text{greedy})<0.3$, $p(\text{selected})>0.8$, and the gap in $p$ is the largest; dagger: chosen without the two thresholds on $p$, by the rule of \cref{fig:gal_beyond}b; items shown elsewhere in the paper are excluded, and each item is shown once. Dashed outline: the target region (specks under 1\,pt not outlined); orange ring: greedy's point; blue ring: \method's point; white rings: other sampled points (merged on the 1\% grid), drawn where they fit and overlap no other ring. Images are cropped to the panel's shape around the target region and the two rings, to at most four times the target region's extent where these fit; (a), (d) and (i) have white margins; instructions are shown without VABench-Point's preamble and output request and without RoboSpatial-Home's opening sentence naming the object. Set names as in \cref{tab:app_embodied}.}
\label{fig:gal_robot}
\end{figure}

\begin{figure}[htp]
\centering
\includegraphics[width=\linewidth]{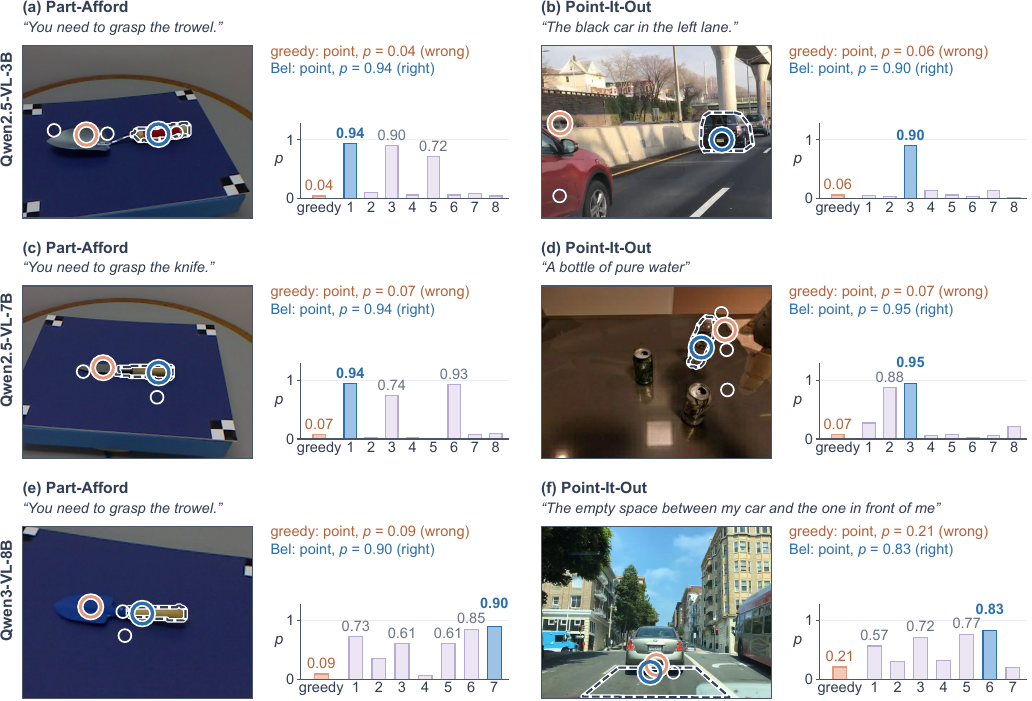}
\caption{\textbf{On Part-Afford and Point-It-Out too, for each of the three policies, an item on which \method moves a wrong greedy point onto the target.} Items chosen by the rule of \cref{fig:gal_robot}; all six meet both thresholds on $p$. Bars: every distinct candidate's $p(\text{correct})$ in sampling order, with the values of greedy, of \method's choice and of every bar at 0.5 or above. Part-Afford's instructions are shown without the request sentence of its prompt; markings and crops as in \cref{fig:gal_robot}.}
\label{fig:gal_robot_new}
\end{figure}

\paragraph{Pooled results.}
\Cref{tab:app_embodied} gives the five sets pooled.
Pooled over the five sets, \method's gains over greedy decoding and over self-consistency are significant for the two Qwen2.5-VL policies and for Qwen3-VL-235B-A22B (\cref{sec:app_q235}), and its gain over the judge for all four policies (95\% intervals over items: over self-consistency [+6.6, +12.7], [+8.2, +14.5] and [+2.1, +7.3] for Qwen2.5-VL-3B, Qwen2.5-VL-7B and Qwen3-VL-235B-A22B; over the judge [+2.8, +8.7], [+7.3, +14.0], [+1.9, +8.0] and [+5.3, +11.2] for Qwen2.5-VL-3B, Qwen2.5-VL-7B, Qwen3-VL-8B and Qwen3-VL-235B-A22B). For Qwen3-VL-8B, \method is significantly ahead of majority vote ([+0.7, +6.5]) and the judge on the five sets (\cref{tab:app_embodied}), and on the 2,346 items of Part-Afford and Point-It-Out it gains 3.3 points over greedy decoding, significantly on both benchmarks (\cref{tab:app_embodied_new}).

\paragraph{Two more benchmarks.}
Part-Afford \cite{yuan2026embodiedr1}, Embodied-R1's grasp-part pointing benchmark on tool images of the RGB-D Part Affordance dataset \cite{myers2015affordance}, has 2,000 items, each with the graspable part as the target.
Point-It-Out \cite{xue2025pointitout} takes its household scenes from Where2Place, already one of our sets, and its kitchen, driving and robot-manipulation scenes from other sources; we take the 346 items of its referred-object localization and task-driven pointing stages on the latter.
The instruction is Part-Afford's evaluation prompt (``Provide one or more points coordinate of objects region this sentence describes: \ldots'') or the Point-It-Out item's own instruction, followed by the same output request; candidates, selectors and thresholds are as above (\cref{tab:app_embodied_new}), as is the rule that an answer's first point is the action.
Each item is again scored on one point (Part-Afford's scoring script averages over all predicted points, and Point-It-Out scores these stages with a normalized box IoU).
Pooled over the 2,346 items, \method's gains over greedy decoding (17.8, 13.2 and 3.3 points for Qwen2.5-VL-3B, Qwen2.5-VL-7B and Qwen3-VL-8B), majority vote, self-consistency and the judge are significant for these three policies (95\% intervals over items for Qwen2.5-VL-3B, Qwen2.5-VL-7B and Qwen3-VL-8B: over majority vote [+15.5, +19.5], [+10.8, +14.5] and [+3.5, +6.9]; over self-consistency [+16.8, +20.9], [+15.3, +19.3] and [+3.5, +6.8]; over the judge [+10.8, +14.8], [+14.2, +18.2] and [+7.5, +11.5]).

\paragraph{Examples.}
\Cref{fig:gal_robot,fig:gal_robot_new} show one item for each policy and pointing set, chosen by a rule: among the items on which \method, at the thresholds of \cref{tab:app_embodied} ($\tau{=}0.8$, $m{=}0$), replaced a greedy point off the target region by a sampled point on it and the decision model gave greedy's point less than 0.3 and the selected point more than 0.8, the item with the largest gap in $p$; for a policy and set without such an item, the repaired item with the largest gap without the two thresholds on $p$, the rule of \cref{fig:gal_beyond}b, is shown and marked with a dagger.
The items of \cref{fig:embodied} and of \cref{fig:gal_beyond} are excluded, and each item is shown once.

\clearpage
\subsection{Variants of the Override Rule}
\label{sec:app_ablation}

For GUI-R1-3B (evaluation subsets), thresholds cross-fitted on each held-out test set give 63.2, 44.2, 75.1, 78.1 and 79.8\% on AC-Low, GUI-Odyssey, GUI-Act-Web, OmniAct-Web and OmniAct-Desktop, against 63.2, 44.4, 75.2, 78.6 and 81.4\% with the development-set thresholds: never higher.
Two further variants coincide with the highest score, the row ``\method w/o override rule'' of \cref{fig:ablation}: adding the vote share to the score selects the same actions on every set, and preferring greedy's action type selects the same actions on the held-out sets, because the setting chosen on the whole development set turns the preference off.

\clearpage
\subsection{Judging Single GUI Steps on the Screen}
\label{sec:app_scorers}

\begin{figure}[htp]
\centering
\includegraphics[width=\linewidth]{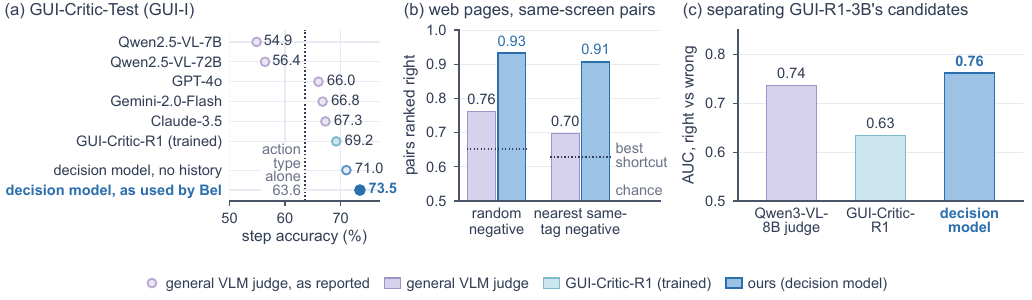}
\caption{\textbf{The decision model is the most accurate step judge on GUI-Critic-Test (GUI-I), ranks the right element first far more often than a prompted judge on Multimodal-Mind2Web, and separates GUI-R1-3B's candidates significantly better than the Qwen3-VL-8B judge and GUI-Critic-R1 on average over the six step-level sets.} (a) Step accuracy on GUI-Critic-Test (GUI-I split), the reported judges plus the decision model with and without the action history; other judges as reported by \cite{wanyan2025guicritic} (lilac and teal; weaker open models omitted); dotted: predicting from the action type alone. (b) Same-screen pairs on Multimodal-Mind2Web: share of pairs in which the right element scores higher than a random or the nearest same-tag element (ties count half); general VLM judge: Claude Haiku 4.5; dotted: the best rule using only the element's position or size. (c) AUC for right vs wrong candidates of GUI-R1-3B, mean over these six sets (evaluation subsets); general VLM judge: Qwen3-VL-8B; both differences significant (paired bootstrap over tasks, 95\%). Higher is better.}
\label{fig:scorers}
\end{figure}

\Cref{fig:scorers} scores actions one at a time, without choosing among them: single steps of GUI-Critic-Test, same-screen element pairs of Multimodal-Mind2Web (construction: \cref{sec:app_prompts}), and right versus wrong candidates of GUI-R1-3B on the six step-level test sets.
\Cref{fig:scorers}a gives step accuracy on the GUI-I split of GUI-Critic-Test; without the screenshot, the decision model's accuracy there falls from 73.5\% to 57.6\%, below the 63.6\% of a rule that reads only the action type, and without the action history it is 71.0\%.
The six questions were written for this evaluation and are used unchanged in every other experiment. The figure reports the verdict, the answer chosen as the score on AC-High (\cref{sec:method}), taken at its more likely answer, with no threshold fitted on GUI-Critic-Test. On GUI-I, the holistic answer (``Is this decision correct?'') reaches 59.9\% and the per-step minimum of the other four answers (\cref{sec:app_prompts}) 56.7\%, both thresholded at 0.5.
On GUI-W (428 steps in the released data), the decision model's step accuracy is 81.8\% (70.8\% without the screenshot), above every judge reported by \cite{wanyan2025guicritic}, including GUI-Critic-R1 (63.1\%) and the proprietary models (62.9 to 68.5\%).

\begin{figure}[htp]
\centering
\includegraphics[width=\linewidth]{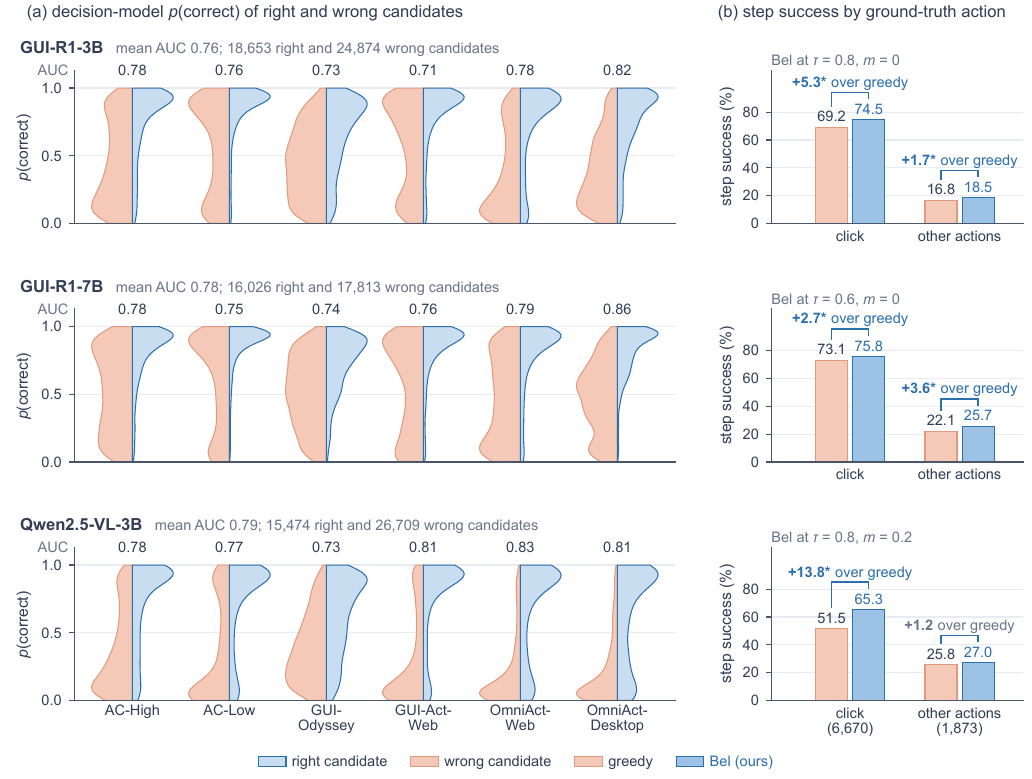}
\caption{\textbf{For every policy, a right candidate usually scores higher than a wrong one on each step-level test set, and \method significantly raises step success on clicks.} One row per policy, GUI-R1-3B as in \cref{fig:analysis}b,c; evaluation subsets of the six step-level test sets. (a) The decision model's $p(\text{correct})$ of the right (blue) and wrong (peach) distinct candidates of every step with more than one, each half scaled to its own maximum; above: AUC of right versus wrong candidates per test set (0.71 to 0.86); beside each policy: its mean over the six sets and the number of candidates. (b) Step success (higher is better) of greedy decoding and of \method at the policy's development setting (\cref{tab:app_thresholds}; AC-High cross-fitted by task, as in \cref{tab:main}), pooled over the six sets, on the steps whose ground truth is a click (6,670) or another scored action (1,873: typing, scrolling, opening an app); labels: gain over greedy decoding (star: 95\% interval excludes zero; paired bootstrap over tasks, each test set resampled separately).}
\label{fig:app_scores_pol}
\end{figure}

\begin{table}[htp]
\centering\small
\caption{\textbf{The decision model ranks a step's candidates significantly better than the Qwen3-VL-8B judge and GUI-Critic-R1, across action types too.} Six step-level sets (evaluation subsets), every distinct candidate scored by all three. Top-1 right: on the steps with a right and a wrong candidate, the top-scored candidate is right (\%; a tie goes to the earlier candidate, greedy's first, as in the override rule); right~$>$~wrong: share of the right/wrong pairs of those steps scored in that order (\%; ties count half), overall and for pairs of different action types; ECE: expected calibration error of the score over every candidate (10 bins). Every difference between the decision model and either judge is significant (paired bootstrap over tasks, 95\%).}
\label{tab:app_rank}
\setlength{\tabcolsep}{2.5pt}
\begin{tabular}{llcccc}
\toprule
Policy & Scorer & Top-1 right $\uparrow$ & Right $>$ wrong $\uparrow$ & Across types $\uparrow$ & ECE $\downarrow$ \\
\midrule
\ours GUI-R1-3B & Decision model (\method) & 67.6 & 71.0 & 72.0 & 0.19 \\
 & Qwen3-VL-8B judge & 62.1 & 65.7 & 68.8 & 0.35 \\
 & GUI-Critic-R1 & 49.0 & 56.0 & 52.4 & 0.41 \\
\midrule
\ours GUI-R1-7B & Decision model (\method) & 66.8 & 69.5 & 77.3 & 0.19 \\
 & Qwen3-VL-8B judge & 63.9 & 66.0 & 74.0 & 0.34 \\
 & GUI-Critic-R1 & 49.6 & 56.7 & 58.1 & 0.40 \\
\midrule
\ours Qwen2.5-VL-3B & Decision model (\method) & 76.3 & 79.1 & 79.8 & 0.15 \\
 & Qwen3-VL-8B judge & 70.6 & 73.3 & 76.6 & 0.35 \\
 & GUI-Critic-R1 & 58.7 & 63.1 & 70.9 & 0.37 \\
\bottomrule
\end{tabular}
\end{table}

\paragraph{Ranking a step's candidates.}
Selection needs the right candidate of a step to score above its wrong ones.
On the six step-level sets, with every distinct candidate scored by the decision model, the Qwen3-VL-8B judge and GUI-Critic-R1, the decision model puts a right candidate first more often than either judge for every policy, orders right above wrong candidates more often, also when the two are of different action types, and its scores are better calibrated (\cref{tab:app_rank}); every one of these differences is significant.

\paragraph{Scores for every policy.}
\Cref{fig:app_scores_pol} repeats \cref{fig:analysis}b,c for each policy.
The decision model separates the right and wrong candidates of GUI-R1-7B and Qwen2.5-VL-3B with an AUC of 0.73 to 0.86 on every step-level test set (mean over the six sets 0.78 and 0.79; GUI-R1-3B: 0.76), and \method raises step success on clicks and on other actions for all three policies, significantly on clicks for every policy and on other actions for both GUI-R1 policies.

\clearpage
\subsection{Scaling to Qwen3-VL-235B-A22B}
\label{sec:app_q235}
\begin{table}[htp]
\centering\small
\caption{\textbf{\method also significantly lifts the much larger Qwen3-VL-235B-A22B, served through an API, on every step-level test set and on ScreenSpot.} Step success (SR, \%) on the six GUI test sets (evaluation subsets) and grounding accuracy (\%) on ScreenSpot, with the candidates, prompts and protocol of \cref{tab:main}; \method's thresholds are chosen on this policy's AC-High candidates ($\tau{=}1.01$, $m{=}0.05$). Gain: difference of the printed values, star: its paired 95\% bootstrap interval excludes zero.}
\label{tab:app_q235}
\setlength{\tabcolsep}{4pt}
\resizebox{\textwidth}{!}{%
\begin{tabular}{lcccccccc}
\toprule
 & \multicolumn{7}{c}{Step success (SR) $\uparrow$} & \makebox[0pt]{Grounding $\uparrow$} \\
\cmidrule(lr){2-8}\cmidrule(lr){9-9}
Selector & AC-High (dev) & AC-Low & GUI-Odyssey & GUI-Act-Web & OmniAct-Web & OmniAct-Desk. & Mean & ScreenSpot \\
\midrule
Policy alone (greedy) & 54.0 & 70.3 & 38.3 & 82.2 & 76.2 & 85.9 & 67.8 & 89.7 \\
\ours\quad \method (ours) & 63.9 & 73.1 & 48.1 & 83.6 & 81.2 & 88.1 & 73.0 & 92.0 \\
\ours \gainrow{gain over the policy alone} & {\footnotesize\textcolor{gain}{+9.9\rlap{*}}} & {\footnotesize\textcolor{gain}{+2.8\rlap{*}}} & {\footnotesize\textcolor{gain}{+9.8\rlap{*}}} & {\footnotesize\textcolor{gain}{+1.4\rlap{*}}} & {\footnotesize\textcolor{gain}{+5.0\rlap{*}}} & {\footnotesize\textcolor{gain}{+2.2\rlap{*}}} & {\footnotesize\textcolor{gain}{+5.2\rlap{*}}} & {\footnotesize\textcolor{gain}{+2.3\rlap{*}}} \\
\bottomrule
\end{tabular}}
\end{table}

\paragraph{Setup.}
Qwen3-VL-235B-A22B \cite{qwen3vl}, a mixture-of-experts model with 22B active parameters, is served by AWS Bedrock and gets the prompts of the other policies: GUI-R1's prompts on the GUI test sets, with the reply started with \texttt{<think>} as those prompts ask, and the prompt of \cref{sec:app_embodied} for robot pointing.
It answers on its 0--1000 coordinate grid, which is mapped to the image.
Each step has the greedy answer (temperature 0) and eight samples at temperature 1.0, merged as for the other policies; images are sent at their original size and as stored, except Multimodal-Mind2Web's PNG screenshots, sent as JPEG at quality 95.
The thresholds of \method and of the Qwen3-VL-235B judge are chosen on this policy's AC-High candidates; for robot pointing \method keeps GUI-R1-3B's thresholds, as for the other policies.
GUI-Critic-R1 is not run for this policy, and the Qwen3-VL-8B judge is not run for it on Multimodal-Mind2Web.
For this policy the development set chooses $\tau{=}1.01$, $m{=}0$ for the Qwen3-VL-8B judge and CATTS's threshold $\tau_{\mathrm{C}}{=}0.8$.

\paragraph{Results.}
On the GUI test sets (\cref{tab:app_q235}), \method significantly raises step success on every step-level test set, by 5.2 points on average, and ScreenSpot accuracy by 2.3 points.
On average over the six step-level test sets, it has the highest step success of every selector (73.0\%, against 72.7\% for region voting) and is significantly ahead of majority voting, CATTS, the Qwen3-VL-8B judge and the Qwen3-VL-235B judge, which judges its own model's candidates (71.8, 72.1, 71.9 and 71.5\%).
On Multimodal-Mind2Web (\cref{sec:app_m2w}; mean over the three splits), \method raises step success from 56.9 to 60.9\%, significantly on each split and to the highest value of the selectors run for this policy; it is significantly ahead of CATTS and the Qwen3-VL-235B judge (59.8 and 58.9\%), and majority and region voting reach 60.2 and 60.7\%.
On robot pointing, pooled over the five sets (722 items), \method reaches 60.4\%, against 56.1\% for greedy decoding, 56.0\% for majority vote, 55.7\% for self-consistency and 52.2\% for the Qwen3-VL-8B judge, each significantly lower.

\clearpage
\subsection{Robustness to the Mark and the Thresholds}
\label{sec:app_robust}
\begin{table}[htp]
\begin{minipage}[c]{0.5\linewidth}
\caption{\textbf{With every mark tested, \method keeps a significant gain over greedy decoding.} GUI-R1-3B, \method at its development setting ($\tau{=}0.8$, $m{=}0$, unchanged); SR: mean step success (\%) over the five held-out step-level sets. Default: a red translucent circle of radius 2.5\% of the longer side; half or double its radius and line; its outline without fill; the default in green. Each variant re-scores every click candidate of the steps whose candidates differ in outcome, in new calls. Gain: of the printed values; brackets: paired 95\% interval over tasks.}
\label{tab:app_marks}
\end{minipage}\hfill
\begin{minipage}[c]{0.46\linewidth}
\centering\footnotesize
\setlength{\tabcolsep}{2.5pt}
\begin{tabular}{lcc}
\toprule
Mark & SR $\uparrow$ & Gain over greedy $\uparrow$ \\
\midrule
Policy alone (greedy) & 63.9 & -- \\
Default (\method) & 68.5 & +4.6 [+3.9, +5.5] \\
Half size & 68.4 & +4.5 [+3.7, +5.4] \\
Double size & 68.3 & +4.4 [+3.6, +5.2] \\
Outline only & 68.8 & +4.9 [+4.1, +5.7] \\
Green & 68.6 & +4.7 [+3.9, +5.6] \\
\bottomrule
\end{tabular}
\end{minipage}
\end{table}

\paragraph{The mark.}
\method draws a click as a translucent red circle with a radius of about 2.5\% of the screenshot's longer side (\cref{sec:method}).
\Cref{tab:app_marks} re-scores every click candidate of GUI-R1-3B on the steps whose candidates differ in outcome with four other marks, keeping \method's thresholds and naming each mark in the image label: half or twice the radius and line, an outline without fill, and a green circle.
Every variant keeps a significant gain over greedy decoding, 4.4 to 4.9 points over the five held-out sets: the gain is insensitive to how the mark is drawn, as long as the action is drawn where it acts.

\begin{figure}[htp]
\centering
\includegraphics[width=\linewidth]{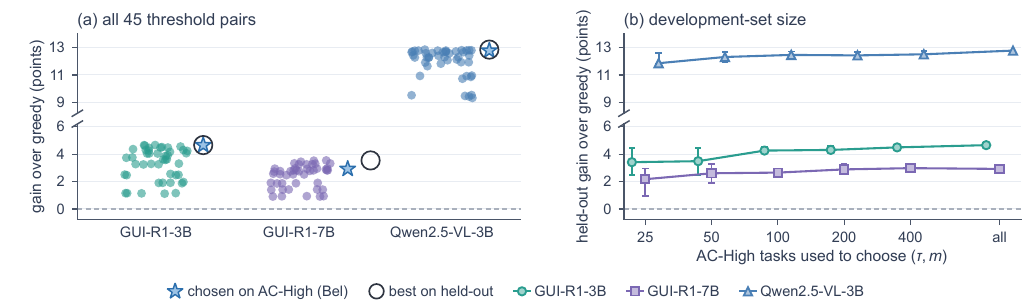}
\caption{\textbf{Every threshold pair of the grid beats greedy decoding, and a small development set suffices.} (a) Gain over greedy decoding (dashed line at zero) in mean step success over the five held-out step-level sets, one dot per $(\tau, m)$ pair of the grid (45 per policy, spread sideways); star: the pair chosen on AC-High, \method's setting; open circle: the best pair on the held-out sets. Held-out means of greedy decoding and of the chosen pair: 63.9 and 68.5 (GUI-R1-3B), 68.3 and 71.2 (GUI-R1-7B), 48.2 and 60.9 (Qwen2.5-VL-3B); 22, 30 and 34 of the 45 pairs are less than 1 point below the chosen one. (b) Held-out gain over greedy decoding of the pair chosen on a random subset of AC-High's tasks (log scale; all: the 1,016 tasks): mean (marker) and interquartile range (bar) of 30 draws per size. Higher is better.}
\label{fig:app_thresh}
\end{figure}

\begin{figure}[htbp]
\centering
\includegraphics[width=\linewidth]{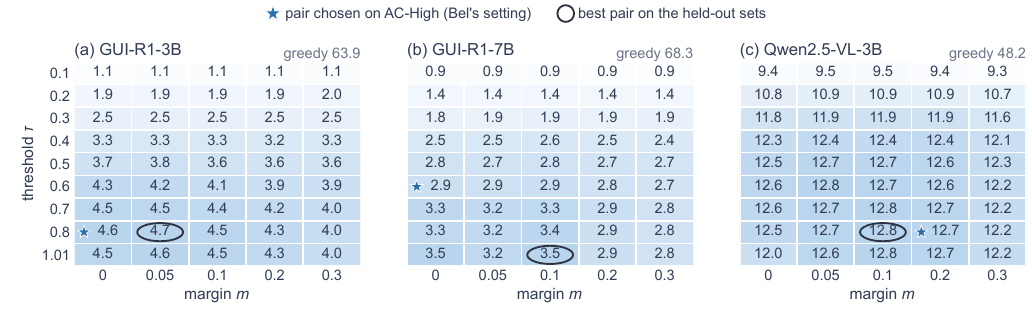}
\caption{\textbf{For every policy, each of the 45 threshold pairs raises held-out step success over greedy decoding, and the gain changes mainly with $\tau$.} Gain over greedy decoding (points) in mean step success over the five held-out step-level test sets (evaluation subsets) for every pair $(\tau, m)$ of the grid: rows $\tau$ ($\tau{=}1.01$ removes the bound on greedy's score), columns $m$; shading scaled to each policy's own range; ``greedy'': greedy decoding's held-out mean. Star: the pair chosen on AC-High, with which \method is reported; ring: the pair with the highest held-out mean. The smallest gains are 1.1, 0.9 and 9.3 points for GUI-R1-3B, GUI-R1-7B and Qwen2.5-VL-3B; within a row (varying $m$) the gain spans 0.3, 0.3 and 0.4 points on average, within a column (varying $\tau$) 3.3, 2.3 and 3.2 points.}
\label{fig:app_thresh_heat}
\end{figure}

\paragraph{The thresholds.}
For each of the three policies, all 45 pairs of the grid of \cref{sec:app_algorithm} give a higher mean step success over the five held-out sets than greedy decoding (\cref{fig:app_thresh}a); the gain depends mainly on $\tau$ and hardly on $m$, and the pair chosen on AC-High is within 0.1, 0.6 and 0.1 points of the best pair for GUI-R1-3B, GUI-R1-7B and Qwen2.5-VL-3B.
Choosing the pair on a random subset of the 1,016 tasks of AC-High's evaluation subset costs little (\cref{fig:app_thresh}b): with 100 tasks the held-out mean is on average 0.3 to 0.4 points below that of the full choice, and no draw of any size falls below greedy decoding.
\Cref{fig:app_thresh_heat} gives the held-out gain of each of the 45 pairs for each policy.

\clearpage
\section{Qualitative Examples}
\label{sec:app_qualitative}

This section states the rules by which the paper's examples are chosen and shows more steps chosen by them.

\subsection{Case Selection Rules}
\label{sec:app_cases}

The steps of \cref{fig:cases} are chosen by a rule among GUI-R1-3B's evaluation steps: \method's choice is right and within 5\% of the screen of the ground-truth point, while the choices of greedy decoding, majority and region voting, CATTS, GUI-Critic-R1 and the Qwen3-VL-8B and Qwen3-VL-235B judges, each at its own development setting (on AC-High cross-fitted), are all wrong; every chosen candidate is a click, and all chosen points and the target lie within 35\% of the screen width and height of each other so that one crop shows them. Steps shown elsewhere in the paper are excluded, as are three desktop steps whose screen already shows the requested page. Per platform (mobile, web, desktop), the step with the largest gap between the decision model's $p$ for \method's choice and its highest $p$ among the other choices is shown, a platform without one being filled by the remaining step with the largest gap; only two steps qualify, both on mobile screens, and the figure shows both.
Steps on which \method is right and every other selector of \cref{tab:main} wrong are 115 of GUI-R1-3B's 9,431 evaluation steps, and the reverse holds on 50; a step counts as right as in step success, and by its action type on the 888 steps that step success leaves out. On the scored steps alone the counts are 86 and 44. On the complete test sets, \method repairs 708, 250, 2,536, 46, 110 and 49 of GUI-R1-3B's steps on AC-High, AC-Low, GUI-Odyssey, GUI-Act-Web, OmniAct-Web and OmniAct-Desktop, 3,699 in all (\cref{fig:qual}a).

The repaired steps shown first in \cref{sec:app_gallery} are chosen by the repair rule: \method switched, greedy was wrong and the selected action right, the decision model gave greedy below 0.3 and the selected action above 0.8 and every wrong candidate at least 0.2 less than the selected action, the two clicks lie within 35\% of the screen width and height of each other so that one crop shows both, and the gap in $p$ is largest; it is applied to GUI-R1-3B's steps of the evaluation subsets on GUI-Odyssey, on the two web test sets together and on OmniAct-Desktop.

The step of \cref{fig:method} is chosen among GUI-R1-3B's AC-High evaluation steps at the development setting by the repair rule with two more conditions: majority voting picks a wrong candidate, and the step has exactly four distinct candidates, all clicks, as the figure's layout needs; steps shown elsewhere in the paper are excluded, and one step qualifies.

The robot-pointing item of \cref{fig:embodied} is chosen among Qwen2.5-VL-7B's items of the five pointing sets at the same setting by the repair rule without the condition on distance: \method switched, greedy was wrong and the selected point right, greedy below 0.3, the selected point above 0.8 and every wrong point at least 0.2 less, and self-consistency (the medoid point) wrong as well; every candidate must be a point with a score, items shown in \cref{sec:app_gallery} are excluded, and the largest gap in $p$ comes first.
The live case of \cref{fig:online_case} is chosen with the same two thresholds on $p$ (greedy below 0.3, the selected action above 0.8) among episodes that succeeded with \method and failed with greedy decoding (same policy, task and task seed), at the first step on which \method changed the action, so that every earlier action is greedy's own, when both candidates are clicks and greedy's action, by our reading of the screen, does not also advance the task; the gap in $p$ is largest over both benchmarks, both policies and all three seeds.

\clearpage
\subsection{Additional Qualitative Examples}
\label{sec:app_gallery}

\begin{figure}[htp]
\centering
\includegraphics[width=\linewidth]{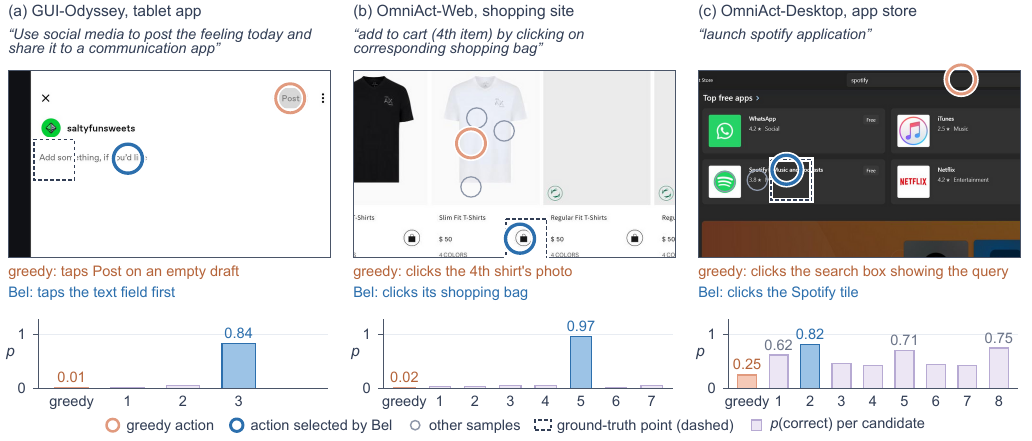}
\caption{\textbf{\method replaces a wrong greedy action with a right sample on tablet, web and desktop screens.} GUI-R1-3B, steps chosen by the repair rule of \cref{sec:app_cases}. Rings: click candidates within the crop that do not overlap another ring; dashed box: the ground-truth point (a click within 14\% of the screen of it is right); bars: every distinct candidate's $p(\text{correct})$ in sampling order.}
\label{fig:gal_cases}
\end{figure}

\begin{figure}[htp]
\centering
\includegraphics[width=\linewidth]{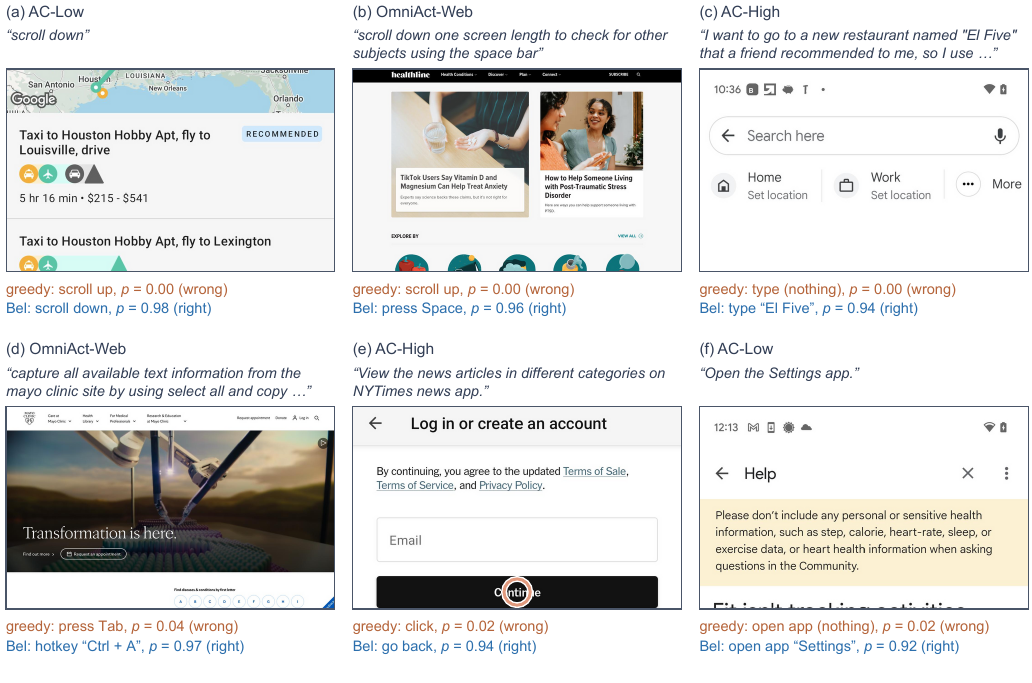}
\caption{\textbf{With the same question, \method also repairs non-click steps.} Repaired steps on which \method replaced a wrong greedy action by a right non-click action (GUI-R1-3B, development setting; on AC-High, the two-fold cross-fitted settings); phone screens are cropped to a band around the positions the actions carry; orange ring: a greedy click; (a)--(c): \cref{fig:nonclick}.}
\label{fig:gal_nonclick}
\end{figure}

\begin{figure}[htp]
\centering
\includegraphics[width=\linewidth]{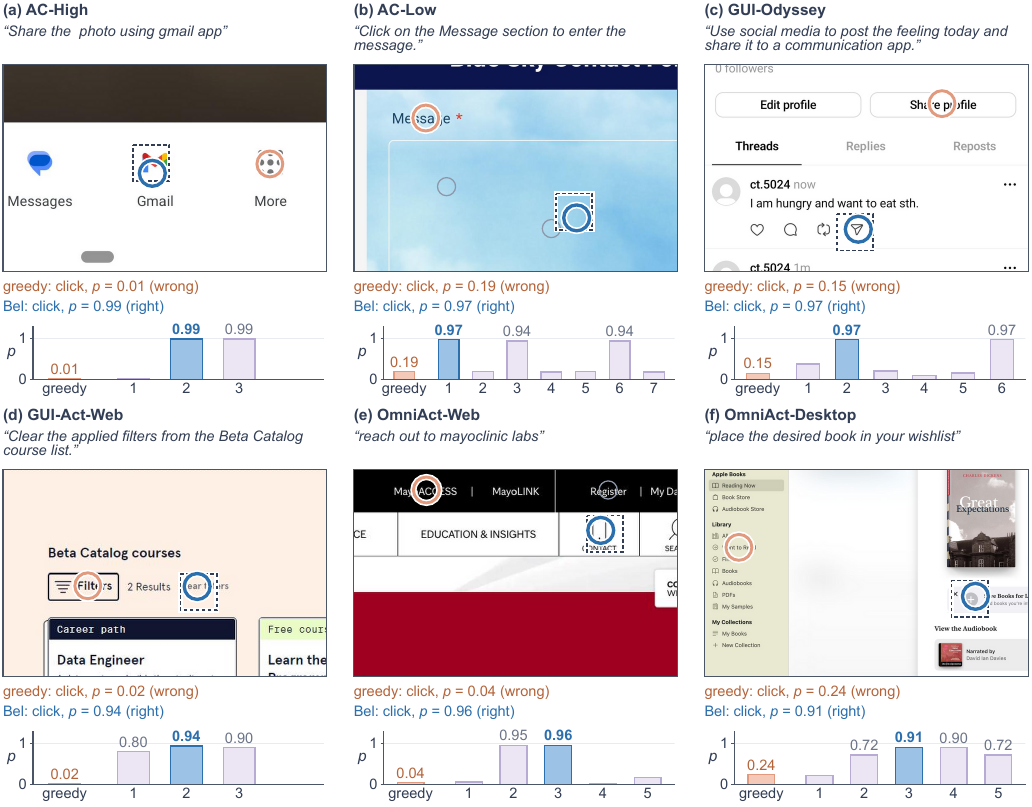}
\caption{\textbf{\method repairs a wrong greedy click on every test set.} One repaired step per test set (GUI-R1-3B, development setting; on AC-High, the two-fold cross-fitted settings): the step not shown in \cref{fig:gal_cases} with the largest gap in $p$ on which \method replaced a wrong greedy click by a right sample. Orange ring: greedy; blue ring: \method's action; grey rings: other click candidates not on another ring; dashed box: ground truth (a click counts as right within 14\% of the screen of it); bars: every distinct candidate's $p(\text{correct})$ in sampling order.}
\label{fig:gal_rep}
\end{figure}

\begin{figure}[htbp]
\centering
\includegraphics[width=\linewidth]{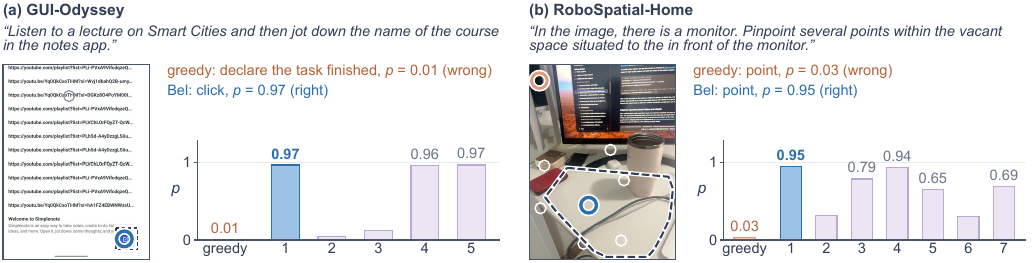}
\caption{\textbf{Beyond clicks and screens, \method keeps a task going that greedy ends too early and moves a robot's point onto its target.} (a) Greedy declares the task finished, an action without a position, while the note is still to be written; \method selects the sample that opens a new note (GUI-R1-3B). (b) A robot pointing item (Qwen2.5-VL-7B): the target region is outlined (dashed) and the other sampled points are white rings; greedy points beside the monitor, \method selects a point in the vacant space in front of it. Markings as in \cref{fig:gal_rep}.}
\label{fig:gal_beyond}
\end{figure}

\begin{figure}[htp]
\centering
\includegraphics[width=\linewidth]{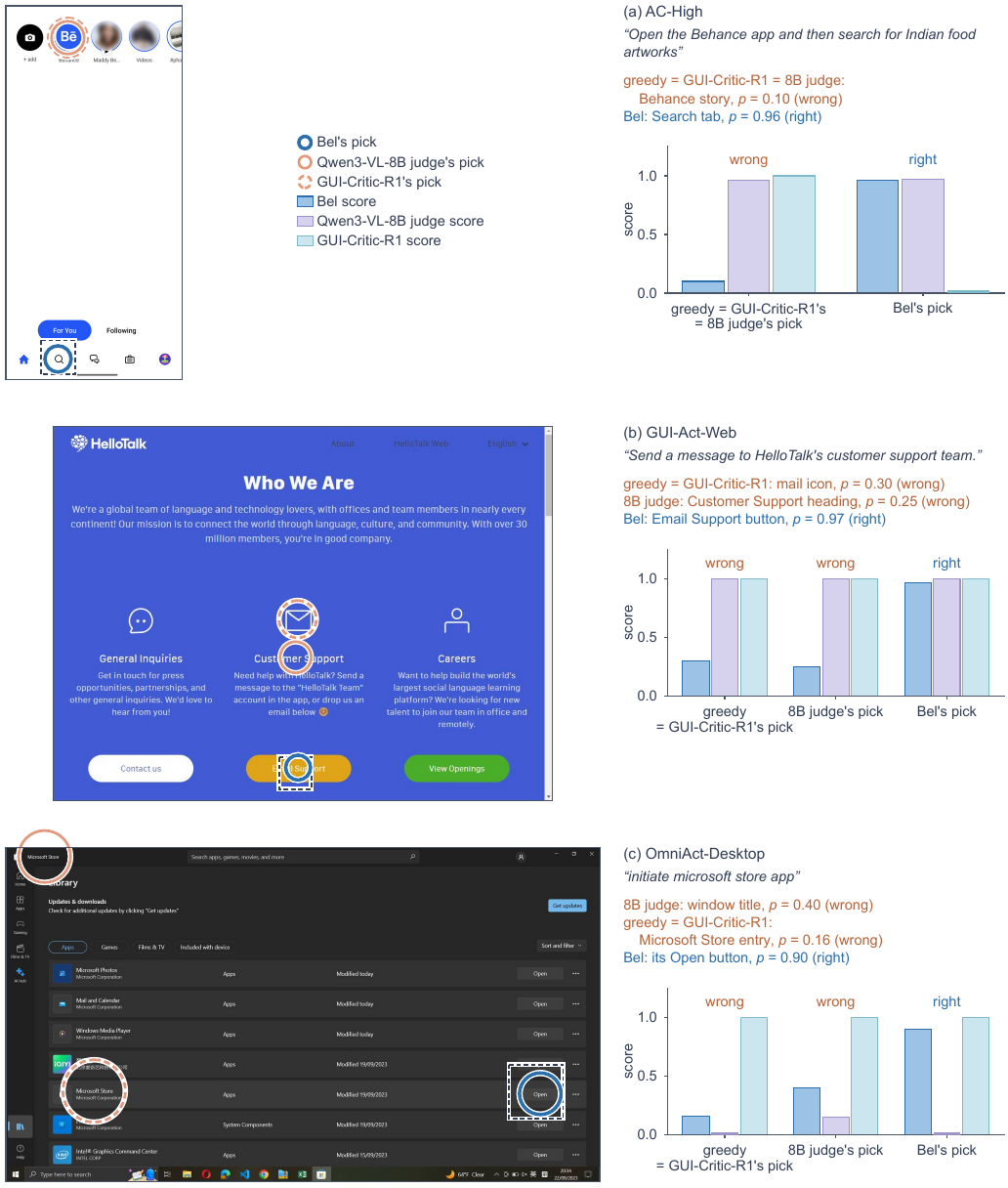}
\caption{\textbf{On these steps \method executes the right action, while the Qwen3-VL-8B judge and GUI-Critic-R1 both execute a wrong one.} GUI-R1-3B, every selector at its own development setting (on AC-High, its two-fold cross-fitted settings, as in \cref{tab:main}). Rings: the action each selector executes (GUI-Critic-R1 keeps greedy's in all three, as does the Qwen3-VL-235B judge, not drawn); dashed box: the ground-truth point; bars: each scorer's score for the candidates the three selectors execute (one group where they coincide).}
\label{fig:gal_judges}
\end{figure}

\Crefrange{fig:gal_cases}{fig:gal_judges} show more steps, all chosen by such rules, none shown elsewhere in the paper except \cref{fig:gal_nonclick}(a)--(c): GUI-R1-3B on the evaluation subsets at its development setting ($\tau{=}0.8$, $m{=}0$; for the AndroidControl-High steps of \cref{fig:gal_nonclick,fig:gal_rep,fig:gal_judges}, the two-fold cross-fitted settings with which AC-High is reported in \cref{tab:main}), and Qwen2.5-VL-7B on the five pointing sets at the same setting.
\Cref{fig:gal_cases} gives the step chosen by the repair rule of \cref{sec:app_cases} on GUI-Odyssey, on the two web test sets together and on OmniAct-Desktop.
\Cref{fig:gal_nonclick} shows that the same question serves actions without a position: for the six kinds of non-click action other than waiting with the largest gaps, the repaired step with the largest gap in $p$ on which \method replaced a wrong greedy action by a right one.
\Cref{fig:gal_rep} gives, for each test set, the repaired step with the largest gap in $p$ under the repair rule of \cref{sec:app_cases}, without its condition on the other wrong candidates.
\Cref{fig:gal_beyond} gives the repaired step with the largest gap among those on which greedy declared the task finished too early, over the six test sets, and the robot pointing item with the largest gap among those on which \method replaced a wrong point by a right one.
\Cref{fig:gal_judges} compares the selectors on the same step: among the steps on which \method's choice is right, within 5\% of the screen of the ground-truth point, the choices of both the Qwen3-VL-8B judge and GUI-Critic-R1 are wrong, at least 20\% away, and the decision model scores \method's choice at least 0.3 above both of theirs, and all four actions (greedy's, \method's and both judges') are clicks; per platform, the step with the largest gap between the decision model's $p$ for \method's choice and for greedy's is shown (\method right and both judges wrong: 269 of GUI-R1-3B's 9,431 evaluation steps; the reverse: 120). GUI-Critic-R1 scores nearly every candidate of these steps close to one and so keeps greedy's action, and the judge scores a wrong candidate about as high as the right one or higher, whereas the decision model tells them apart.

\begin{figure}[htp]
\centering
\includegraphics[width=\linewidth]{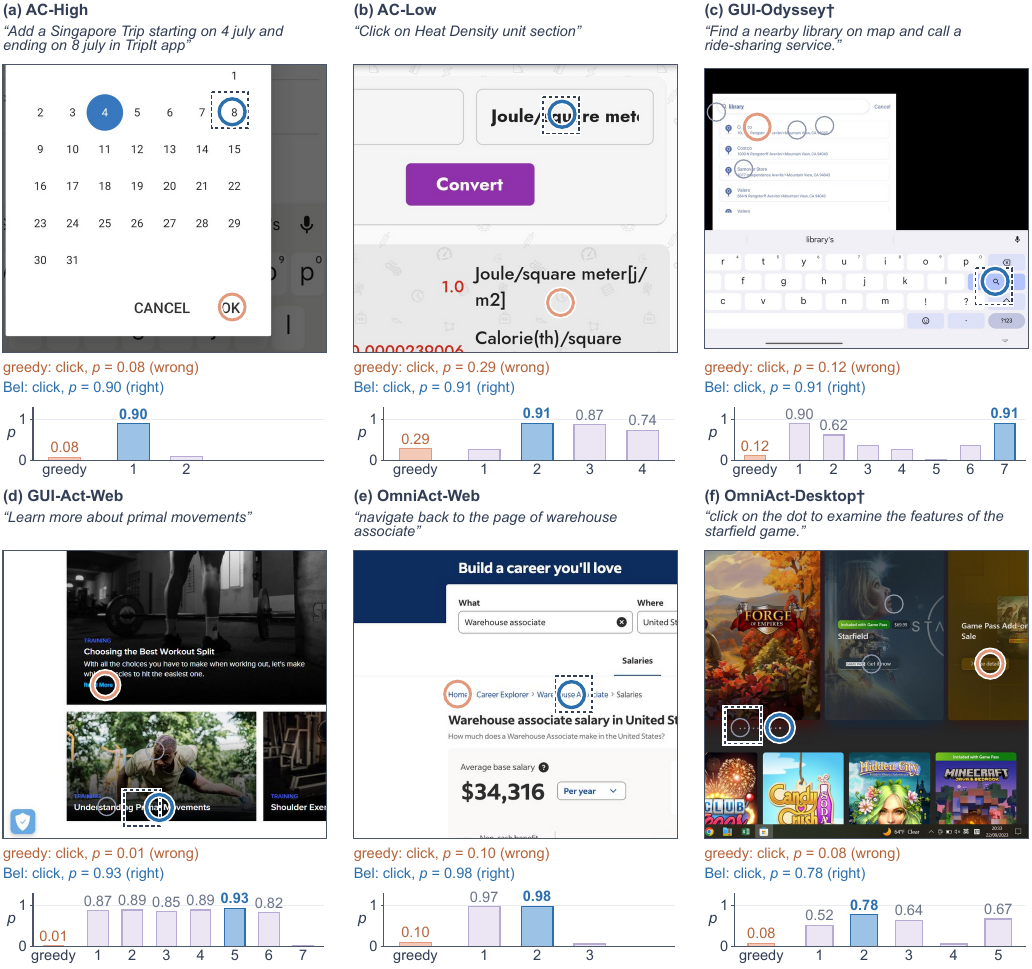}
\caption{\textbf{\method repairs a wrong greedy click of GUI-R1-7B on every test set as well.} Repaired steps of GUI-R1-7B on the evaluation subsets at this policy's development setting ($\tau{=}0.6$, $m{=}0$; on AC-High, the two-fold cross-fitted settings), chosen by the rule of \cref{fig:gal_rep} (\method switched, greedy's click was wrong and the selected click right, the decision model gave greedy below 0.3 and the selected click above 0.8, the two clicks lie within 35\% of the screen width and height of each other, and per test set the step with the largest gap in $p$ is shown) with these additions: both clicks lie on the screen; a step shown in another figure of the paper, or sharing its screenshot or instruction with one, is excluded, so a step is shown for one policy only (policies in the order Qwen3-VL-235B-A22B, GUI-R1-7B, Qwen2.5-VL-3B, relaxed choices after all unrelaxed ones); ties in the gap go to the lower $p$ for greedy. On a test set without such a step ($\dagger$), the rule, its additions kept, is relaxed until a step qualifies: first without the 35\% condition (the crop is enlarged to show both clicks), then with 0.5 in place of the thresholds 0.3 and 0.8 (greedy below, the selected click above), then without thresholds on $p$. Here (c) is from the first relaxation and (f) from the second. Markings as in \cref{fig:gal_rep}; rings: click candidates within the crop that do not overlap another ring.}
\label{fig:gal_pol_guir1-7b}
\end{figure}

\begin{figure}[htp]
\centering
\includegraphics[width=0.985\linewidth]{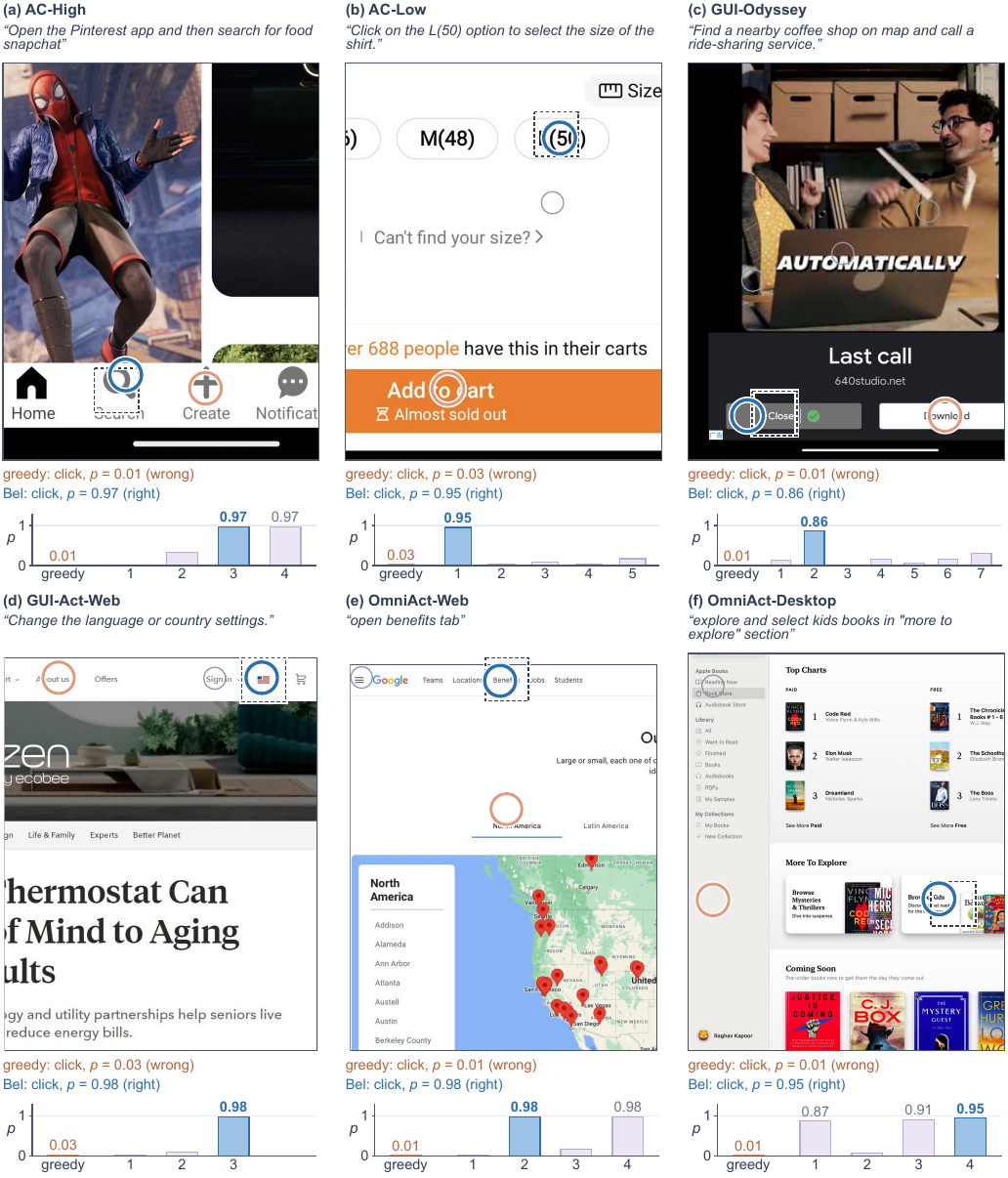}
\caption{\textbf{\method repairs a wrong greedy click on every test set for Qwen2.5-VL-3B as well.} Repaired steps of Qwen2.5-VL-3B on the evaluation subsets at this policy's development setting ($\tau{=}0.8$, $m{=}0.2$; on AC-High, the two-fold cross-fitted settings), chosen by the rule of \cref{fig:gal_pol_guir1-7b}, unrelaxed on every test set: per test set, the step with the largest gap in $p$ on which \method replaced a wrong greedy click by a right sample. Markings as in \cref{fig:gal_rep}, drawn whole where they cross the screen's edge; rings: click candidates within the crop that do not overlap another ring.}
\label{fig:gal_pol_base3b}
\end{figure}

\begin{figure}[htp]
\centering
\includegraphics[width=\linewidth]{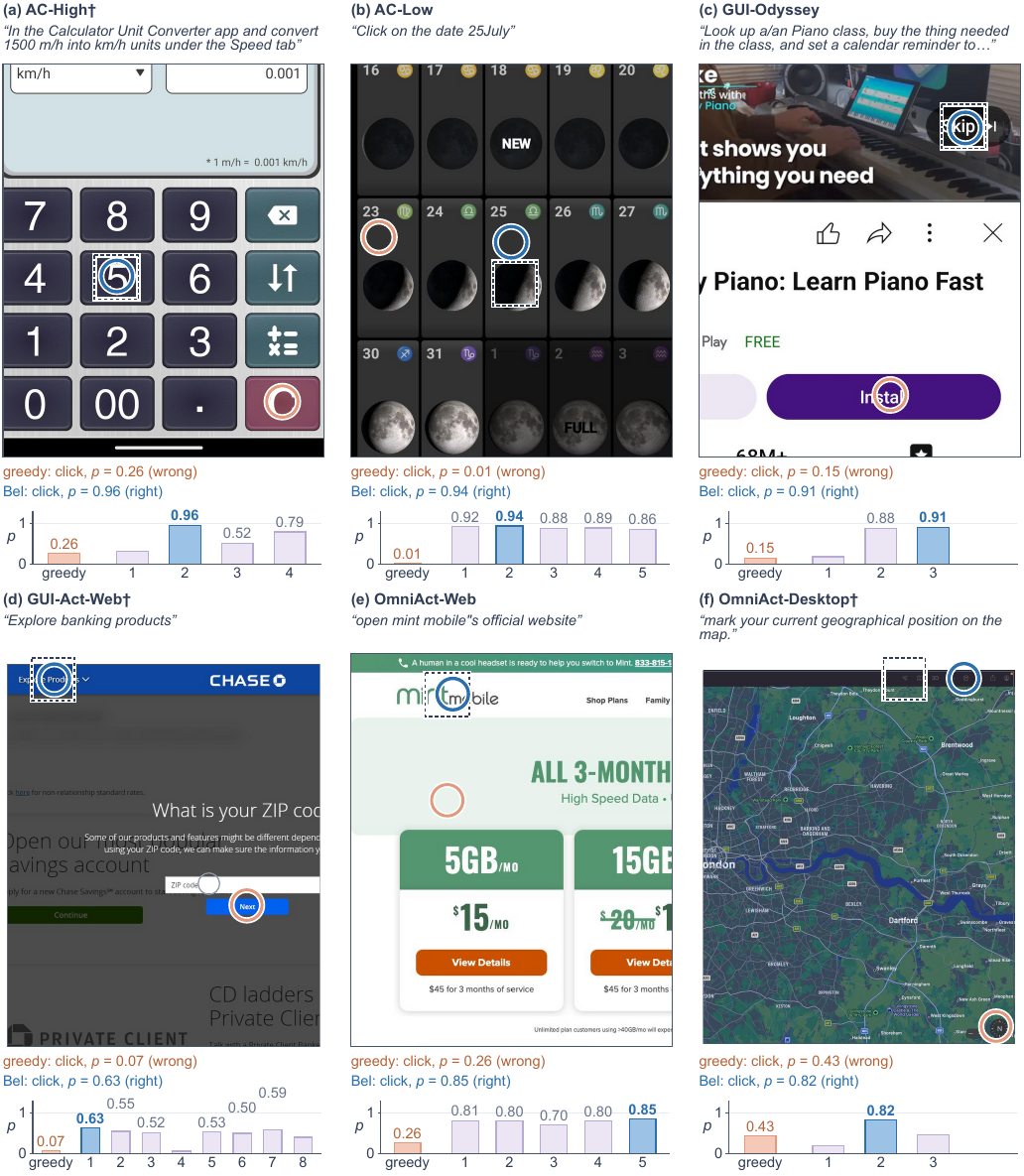}
\caption{\textbf{\method also repairs a wrong greedy click of the much larger Qwen3-VL-235B-A22B on every test set.} Repaired steps of Qwen3-VL-235B-A22B, served through an API, on the evaluation subsets at this policy's development setting ($\tau{=}1.01$, $m{=}0.05$; on AC-High, the two-fold cross-fitted settings), chosen by the rule of \cref{fig:gal_pol_guir1-7b}: per test set, the step with the largest gap in $p$ on which \method replaced a wrong greedy click by a right sample; $\dagger$: relaxed as stated there, (a) at the first relaxation, (d) and (f) at the second. Markings as in \cref{fig:gal_rep}, drawn whole where they cross the screen's edge; rings: click candidates within the crop that do not overlap another ring.}
\label{fig:gal_pol_q235}
\end{figure}

\Crefrange{fig:gal_pol_guir1-7b}{fig:gal_pol_q235} apply the rule of \cref{fig:gal_rep}, with the additions and the relaxation stated in the caption of \cref{fig:gal_pol_guir1-7b}, to GUI-R1-7B, Qwen2.5-VL-3B and Qwen3-VL-235B-A22B, each at its own development setting, and show one repaired step for every policy and test set.

\end{document}